\documentclass{ieeeaccess}
\usepackage{cite}
\usepackage{amsmath,amssymb,amsfonts}
\usepackage{algorithm, algpseudocode}
\usepackage{graphicx}
\usepackage{textcomp}
\usepackage{varwidth}

\usepackage[nolist]{acronym}
\usepackage{upgreek}
\usepackage{mathtools}
\usepackage{romannum}

\usepackage{soul}
\DeclareRobustCommand{\hlnew}[1]{{\sethlcolor{white}\hl{#1}}}

\usepackage{mdframed}

\newacro{lfd}[LfD]{Learning from Demonstration}
\newacro{rm}[RM]{Riemannian Manifold}
\newacro{uq}[UQ]{Unit Quaternion}
\newacro{rl}[RL]{Reinforcement Learning}
\newacro{grl}[$\mathcal{G}$-RL]{Geometric Reinforcement Learning}
\newacro{irl}[IRL]{Inverse Reinforcement Learning}
\newacro{sac}[SAC]{Soft Actor-Critic}
\newacro{ppo}[PPO]{Proximal Policy Optimization}
\newacro{ddpg}[DDPG]{Deep Deterministic Policy Gradient}
\newacro{td3}[TD3]{Twin Delayed Deep Deterministic Policy Gradients}
\newacro{promps}[ProMPs]{Probabilistic Movement Primitives}
\newacro{lqr}[LQR]{Linear Quadratic Regulator}
\newacro{power}[PoWER]{Policy learning by Weighting Exploration with the Returns}
\newacro{blackdrops}[Black-DROPS]{Black-box Data-efficient RObot Policy Search}
\newacro{cmaes}[CMA-ES]{Covariance Matrix Adaptation Evolution Strategy}
\newacro{pvt}[PVT]{Position, Velocity, Time}
\newacro{cobot}[cobot]{Collaborative Robot}
\newacro{bbo}[BBO]{Black Box Optimization}
\newacro{gpp}[GPP]{Gaussian Policy Parameterization}
\newacro{tsgpp}[TSGPP]{Tangent Space Gaussian Policy Parameterization}
\newacro{bpp}[BPP]{Bingham Policy Parameterization}
\newacro{bo}[BO]{Bayesian Optimization}
\newacro{gp}[GP]{Gaussian process}
\newacro{ilqr}[iLQR]{iterative \ac{lqr}}
\newacro{vae}[VAE]{variational autoencoder}
\newacro{spd}[SPD]{Symmetric Positive Definite}
\newacro{dmp}[DMP]{Dynamic Movement Primitives}

\newcommand{\bm}[1]{\boldsymbol{\mathbf{#1}}}

\usepackage{tabularx}
\makeatletter
\newcommand{\multiline}[1]{%
	\begin{tabularx}{\dimexpr\linewidth-\ALG@thistlm}[t]{@{}X@{}}
		#1
	\end{tabularx}
}
\def\algbackskip{\hskip-\ALG@thistlm}
\makeatother

\newcommand{\trsp}{{^{\top}}}

\newcommand{\figref}[1]{Fig.~\hyperref[#1]{\ref*{#1}}}
\newcommand{\figsref}[1]{Figures~\hyperref[#1]{\ref*{#1}}}
\newcommand{\Figref}[1]{Figure~\hyperref[#1]{\ref*{#1}}}

\newcommand{\tabref}[1]{Tab.~\hyperref[#1]{\ref*{#1}}}
\newcommand{\secref}[1]{Section~\hyperref[#1]{\ref*{#1}}}
\newcommand{\algoref}[1]{Algorithm~\hyperref[#1]{\ref*{#1}}}

\newcommand{\eg} {{e.g.,}~} %
\newcommand{\ie} {{i.e.,}~} %
\newcommand{\etal}{\MakeLowercase{\textit{et al.\ }}}

\newcommand\manS[1]{\mathcal{S}^{#1}}
\newcommand\manR[1]{\mathcal{R}^{#1}}

\newcommand\manSO[1]{\mathcal{SO}\left({#1}\right)}
\newcommand\manSPD[1]{\mathcal{S}^{#1}_{++}}

\newcommand*\dashline{\rotatebox[origin=c]{90}{\,-\,-}}

\graphicspath{{figs/}}

\def\BibTeX{{\rm B\kern-.05em{\sc i\kern-.025em b}\kern-.08em
		T\kern-.1667em\lower.7ex\hbox{E}\kern-.125emX}}
\begin{document}
\pagenumbering{arabic}
	\history{Date of publication xxxx 00, 0000, date of current version xxxx 00, 0000.}
	\doi{10.1109/ACCESS.2017.DOI}
	\title{Geometric Reinforcement Learning For Robotic Manipulation}
    
	\author{\uppercase{Naseem Alhousani}\authorrefmark{1,3,4},
		\uppercase{Matteo Saveriano}\authorrefmark{2}, \IEEEmembership{Member, IEEE},
		\uppercase{Ibrahim Sevinc}\authorrefmark{4}, \uppercase{Talha Abdulkuddus}\authorrefmark{3},
		\uppercase{Hatice Kose}\authorrefmark{1}, \IEEEmembership{Member, IEEE}, 
		and 
		\uppercase{Fares J. Abu-Dakka}\authorrefmark{5}, \IEEEmembership{Member, IEEE}}
	\address[1]{Faculty of Computer and Informatics Engineering, Istanbul Technical University, Maslak, 34467 Sar{\i}yer/Istanbul, Turkey (e-mail: nalhousani@itu.edu.tr, hatice.kose@itu.edu.tr)}
	\address[2]{Department of Industrial Engineering (DII), University of Trento, Trento, 38123, Italy (e-mail: matteo.saveriano@unitn.it)}
	\address[3]{ILITRON Enerji ve Bilgi Teknolojileri A.\c{S}, Sultan Selim Mahallesi, Akyol Sanayi Sitesi \c{C}{\i}kmaz{\i} No:10/1 Ka\u{g}{\i}thane Istanbul (e-mail: talha.abdulkuddus@ilitron.com)}
        \address[4]{MCFLY Robot Teknolojileri A.\c{S}, Huzur Mah. Ahmet Bayman Cad. No: 2a Sar{\i}yer / Istanbul, Turkey (e-mail: ibrahim.sevinc@mcflyrobot.com)}
	\address[5]{Technical University of Munich, Germany; Munich Institute of Robotics and Machine Intelligence (MIRMI) (e-mail: fares.abu-dakka@tum.de)}
	\tfootnote{This work has been partially supported by The Scientific and Technological Research Council of Turkey (TÜBİTAK) under Grant 3201141, and by the euROBIN project under grant agreement No. 101070596.}
	
	\markboth
	{Alhousani \headeretal: Geometric Reinforcement Learning}
	{Alhousani \headeretal: Geometric Reinforcement Learning}
	
	\corresp{Corresponding author: Fares~J.~Abu-Dakka (e-mail: fares.abu-dakka@tum.de).}
	
	\begin{abstract}
	    Reinforcement learning (RL) is a popular technique that allows an agent to learn by trial and error while interacting with a dynamic environment. The traditional \ac{rl} approach has been successful in learning and predicting Euclidean robotic manipulation skills such as positions, velocities, and forces. However, in robotics, it is common to encounter non-Euclidean data such as orientation or stiffness, and failing to account for their geometric nature can negatively impact learning accuracy and performance.
		In this paper, to address this challenge, we propose a novel framework for RL that leverages Riemannian geometry, which we call \ac{grl}, to enable agents to learn robotic manipulation skills with non-Euclidean data. Specifically, \ac{grl} utilizes the tangent space in two ways: a tangent space for parameterization and a local tangent space for mapping to a non-Euclidean manifold. The policy is learned in the parameterization tangent space, which remains constant throughout the training. The policy is then transferred to the local tangent space via parallel transport and projected onto the non-Euclidean manifold. The local tangent space changes over time to remain within the neighborhood of the current manifold point, reducing the approximation error. Therefore, by introducing a geometrically grounded pre- and post-processing step into the traditional RL pipeline, our \ac{grl} framework enables several model-free algorithms designed for Euclidean space to learn from non-Euclidean data without modifications. 
        Experimental results, obtained both in simulation and on a real robot, support our hypothesis that \ac{grl} is more accurate and converges to a better solution than approximating non-Euclidean data.
	\end{abstract}
	
	\begin{keywords}
		Learning on manifolds, policy optimization, policy search, geometric reinforcement learning.
	\end{keywords}
	
	\titlepgskip=-15pt
	
	\maketitle
	
\section{Introduction}
	\label{sec:introduction}	
    \begin{figure*}[!t]
    \centering
	\def\svgwidth{\linewidth}
	{\fontsize{8}{8}
		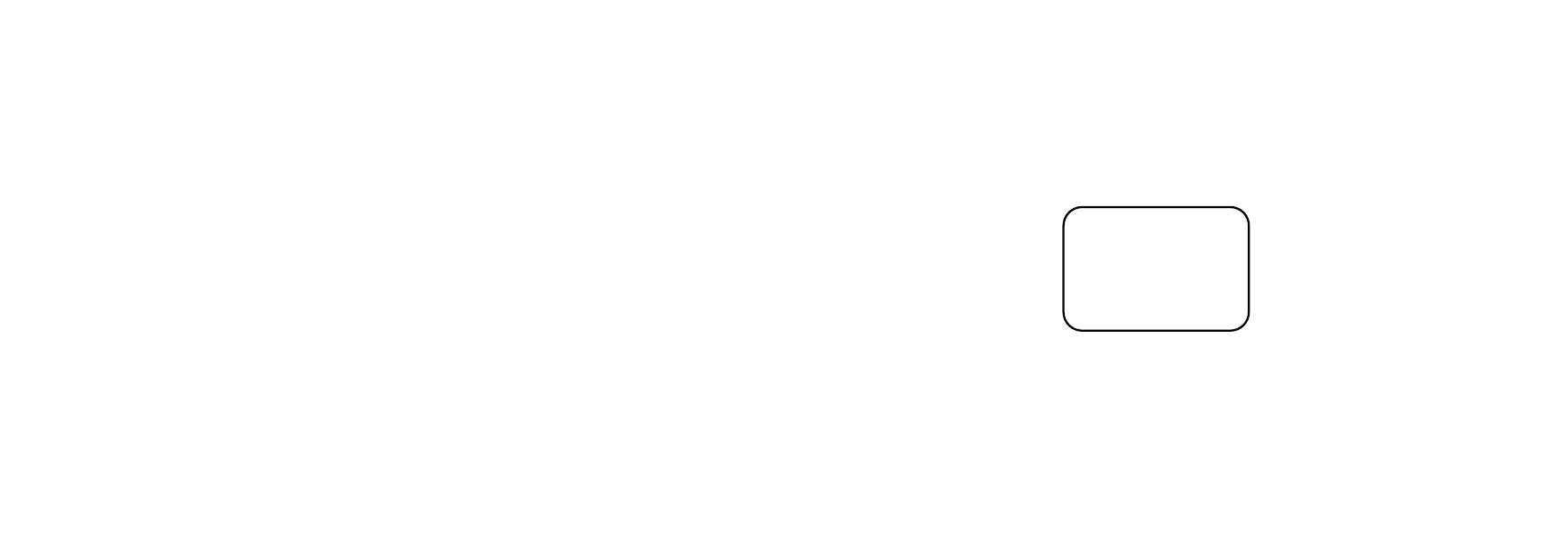}
	\caption{Overview of the proposed framework for \acf{grl}. Starting with the action from the \ac{rl} algorithm, the tangent space vector $\boldsymbol{\mathfrak{a}}_{P}$ is transferred from the parameterization tangent space to the local tangent space through parallel transport. The vector is then mapped onto the corresponding composite manifold to produce the desired action (\eg orientation and impedance), then passed to the controller to execute the action. The new state and the corresponding action are sent to the reward function to evaluate the quality of the current policy. This evaluation is delivered back to the \ac{rl} algorithm, along with the full observed state.}
	\label{fig:flow}
    \end{figure*}
	
	\IEEEPARstart{N}{on-Euclidean} data, like orientation, stiffness, or manipulability, are important in the field of robotics, as they are widely used during learning and implementation processes~\cite{abu2021probabilistic}. \hlnew{To illustrate, consider real-world scenarios in robotics, like assembly tasks, polishing and grinding, and the automation of industrial welding processes. In such contexts, acquiring knowledge of non-Euclidean data, such as orientation and impedance information, becomes pivotal.} Such data have special properties that do not allow the use of Euclidean calculus and algebra. Despite this, they are usually treated as Euclidean data, which demands pre- or post-processing (e.g., normalizing orientation data) to conform to their non-Euclidean nature. This process involves an approximation, and with repetition, the approximation errors will accumulate until it reaches a level that affects the learning process in terms of accuracy and speed to reach the desired results. This issue was noticed early in the field of statistical learning~\cite{pennec2006intrinsic}, as determining the center of a non-Euclidean geometric data set using normalization leads to an error in determining the mean. In robotic manipulation, some data belong to different Riemannian manifolds (\eg 3-sphere manifold $\manS 3$ where the unit quaternions live which are possible representations for orientation and \ac{spd} manifold $\manSPD d$ for stiffness and manipulability), and proper mathematical tools need to be developed in order to avoid approximations~\cite{calinon2020gaussians}.

    Nowadays, the practical applications of \ac{rl} have become many and span various fields~\cite{chen2021rdrl,chen2022game,ren2022privacy} including robotics~\cite{kober2013reinforcement}. In \ac{rl}, a policy produces actions based on the current state. When these actions and/or states have a geometric meaning (manifold data) like orientation, stiffness, or manipulability, it requires pre- and post-processing to preserve and benefit from the data geometry; neglecting the underlying constraints of the manifold of these data leads to inaccuracy in both exploration and learning. This is valid for deterministic and probabilistic \ac{rl} approaches. Nevertheless, a distribution is learned for probabilistic \ac{rl} algorithms instead of a single action. This adds more challenges when it samples manifold data from a distribution like the Gaussian distribution, as illustrated in Fig.~\ref{fig:both}~(b). For both cases, benefiting from Riemannian geometry can potentially improve the quality of learned policies.
    
	In this paper, we propose a novel \ac{rl} (\acf{grl}) framework leveraging Riemannian geometry to exploit the geometric structure of the robotic manipulation data. This framework involves applying policy parameterization on the tangent space of a base point on the manifold, followed by using parallel transport to transport the action in a tangent space that moves with the active point. The result is then mapped to its corresponding non-Euclidean manifold. We apply {\ac{grl}} to learn and predict actions, like orientation data represented as unit quaternions or stiffness and manipulability data encapsulated in {\ac{spd}} matrices. The proposed \ac{grl} framework has been applied to extend two prominent deep reinforcement learning algorithms---\ac{sac} \cite{haarnoja2018soft} and \ac{ppo} \cite{schulman2017proximal})---to work with manifold data.
    Furthermore, we have applied it to \ac{cmaes}~\cite{hansen2006cma} (which belongs to the family of \ac{bbo} algorithms). \ac{cmaes} can be used as a policy improvement method like in~\cite{stulp2013policy}. An overview of \ac{grl} is shown in Fig.~\ref{fig:flow}. 
    
    To summarize, our work can be outlined by the following contributions:
    \begin{itemize}
        \item A novel geometry-aware \ac{rl} framework, namely \ac{grl}, that incorporates Riemannian geometry to enable agents to learn robotic manipulation skills with non-Euclidean data. 
        \item Different instantiations of \ac{grl} to extend popular model-free \ac{rl} approaches including: 
        \begin{itemize}
            \item[--] model-free \ac{rl} algorithms (\eg \ac{power}),
            \item[--] model-free deep \ac{rl} algorithms (\eg \ac{sac} and \ac{ppo}), and
            \item[--] \ac{bbo} algorithms (\eg \ac{cmaes}).
        \end{itemize}
        \item Extensive evaluation and experimentation with simulations and a physical robot, with comparisons to different distributions and baselines.
    \end{itemize}

     The rest of the paper is organized as follows. Section \Romannum{2} discusses related work; Section \Romannum{3} provides a short background about \ac{rl} and Riemannian manifolds; Section \Romannum{4} presents our proposed approach; Section \Romannum{5} shows experimental results from both simulation and a physical robot; Section \Romannum{6} discusses the results and the limits of our approach; and we conclude the paper in Section \Romannum{7}.

\section{Related Work}
Most conventional \ac{rl} algorithms that utilize a Gaussian distribution (e.g.,~\cite{theodorou2010generalized,haarnoja2018soft,schulman2017proximal,kober2011policy,chatzilygeroudis2017black,saveriano2017data, lillicrap2015continuous,fujimoto2018addressing}) are not suitable for accurately learning non-Euclidean data. This is because such data resides in a curved space rather than a vector space and therefore requires a  special treatment to avoid approximation errors and to account for its unique properties of that data. This has been approached in different ways.

The Riemannian manifold, Riemannian metric, and tangent space are mathematical concepts utilized in creating geometrical tools for statistics on manifolds as in~\cite{pennec2006intrinsic}, which have been used in different research works. Abu-Dakka \etal \cite{abu2021probabilistic} leveraged Riemannian geometry in the context of learning robotic manipulation skills (e.g., stiffness, manipulability, and covariance) using a kernelized treatment in the tangent space. Huang~\etal~\cite{huang2020toward} proposed adapting learned orientation trajectories to pass through via-points or end-points while also considering the angular velocity. Work done in~\cite{beik2021learning} utilizes a \ac{vae} to learn geodesics on Riemannian manifolds using \ac{lfd}, which generates end-effector pose trajectories able to dynamically avoid obstacles present in the environment. 

In contact-rich manipulation tasks, it is not safe to only use position control. Research in~\cite{chang2022impedance} combines contact \acp{dmp} with \ac{sac} to adapt impedance and learn both linear and orientation stiffness according to a given force and position trajectories, which is then passed into an adaptive admittance controller for robotic manipulation. However, stiffness in their research is represented as a diagonal matrix, whereas our approach can learn the full stiffness matrix. Representing stiffness as a diagonal matrix avoids computational complexity on account of accuracy. But in some cases, such as examining stability properties, it is important to note that the off-diagonal elements of the stiffness matrix can have a direct impact. Disregarding these elements may result in an imprecise assessment of stability~\cite{kao1997robotic}. Additionally, the off-diagonal elements in the stiffness matrix correspond to the interaction between various degrees of freedom. If these elements are ignored by employing a diagonal matrix, it can result in an oversimplified analysis, causing the loss of vital information. Off-diagonal interactions can occur due to physical connections, inter-dependencies among variables, or constraints within the system. Utilizing the full stiffness matrix allows researchers to precisely account for and assess these interactions.

Authors of~\cite{zhang2021learning} also employ a diagonal stiffness matrix in the context of variable impedance, using \ac{irl} to discover the reward function in addition to the policy from an expert demonstration (\ac{lfd}). They proposed that their algorithm can be extended to the full stiffness matrix using Cholesky decomposition. We used Cholesky decomposition as a baseline, and our results show that our algorithm outperforms this baseline.

In \cite{abu2018force}, authors studied the use of \ac{lfd} for force sensing and variable impedance control, with the proposed framework able to use both Cholesky decomposition and Riemannian manifold representations of stiffness. The main difference from our work is that they did not use \ac{rl}; their work was instead based on \ac{lfd}.

Although researchers in \cite{oikawa2020assembly} proposed a method for online selection of non-diagonal stiffness matrices for admittance control using \ac{rl}, they still learn to select from a few previously defined full stiffness matrices. Our algorithm can learn the full stiffness matrix online.

    In the context of image segmentation, authors in~\cite{colutto2009cma} proposed a method for 3D image reconstruction. They achieved this by modifying the original \ac{cmaes} to work on Riemannian manifolds and applying optimization on the tangent space. Unlike our approach of optimizing in the parameter space where geometric data is parameterized using Euclidean parameters, their technique optimizes geometric data directly. In computer vision, parameterization on the tangent space is commonly used to regress rotations with deep learning, as explained in~\cite{chen2022projective}. 

    In reference~\cite{rozo2022orientation}, the utilization of Riemannian manifolds with a solitary optimized tangent space was employed to ensure compliance of parameterization results with manifold geometry. Our work, in contrast, proposes the utilization of two tangent spaces: one for parameterization and another for mapping in the exploration neighborhood. Specifically, our approach maintains proximity of the local tangent space to the active exploration region of the manifold, resulting in more effective utilization of the Riemannian geometry.
 
    The authors of~\cite{wang2022so} proposed a policy equivariant to $\mathcal{SO}(2)$ when the reward and transition functions are invariant to that group. This work is interesting and makes the learning of the elements of $\mathcal{SO}(2)$ faster. However, it does not discuss how to treat orientation data while learning the policy. Our work learns a policy (i.e., orientation) while considering the geometry of manifold data.
 
    Researchers have explored the application of optimization algorithms on Riemannian manifolds. The authors of reference~\cite{jaquier2020bayesian} employed\ac{bo} to optimize policy parameters and introduced geometry-aware kernels. These kernels enable proper measurement of the similarity between Riemannian manifold parameters using \ac{gp}. Another recent work, in~\cite{jaquier2022geometry}, implemented the geometry-aware Riemannian Mat\'ern kernels in the domain of robotics. These investigations consider non-Euclidean manifolds' geometry and propose a geometry-aware framework. Given the advantages of Riemannian geometry in \ac{bo}, we endeavor to exploit it in the realm of \ac{rl}.

    Policy learning in $\mathcal{SE}(3)$ actions is proposed in~\cite{wang2020policy}, achieved by factorizing high dimensional action spaces into several smaller action spaces with progressively augmented state spaces. Each action space is handled by its own neural network. This work is primarily focused on learning poses by imitation of images. A limitation of this work is its use of Euler angles to represent the orientations and being restricted to $\pm30$ degrees rotations out of the plane. The geometry of the orientation data is also not considered, as there is no explanation or discussion about it in the paper.

	Although there are many existing works in the field of \ac{lfd} and supervised learning, Riemannian geometry has not been exploited in \ac{rl}. A recent work, \ac{bpp}~\cite{james2022bingham}, uses the  Bingham distribution as an alternative to the Gaussian distribution for learning orientation policies. This choice was motivated by the argument that unit quaternions can be directly sampled from the Bingham distribution, unlike the Gaussian distribution, where one must use normalization. Nevertheless, authors in~\cite{james2022bingham} reported that as their implementation uses several neural networks, instability in the learning process could occur if erroneous data is sampled from them. Furthermore, our algorithm is not limited to a special distribution such as the Bingham distribution, which is constrained to the sphere manifold. As a result, it can effectively handle data from other types of manifolds, such as $\manSPD d$.
    We experimentally compare the performance of \ac{grl} and \ac{bpp} in Sec.~\ref{subsec:wahba}.
	
\section{Background}
	\subsection{Reinforcement Learning}
	    The general formulation of a typical \ac{rl} problem is about an agent at time $t$ in state $\mathbf{s}_t $ selecting an action $\bm{a}_t $ according to a stochastic policy
	        \begin{equation}
	            \pi_{\bm{\theta}}(\bm{a}|\bm{s}) = \Pr(\bm{a} = \bm{a}_{t}\mid \bm{s} =\bm{s}_{t}),
	        \end{equation}
        where $\bm{\theta}\in \manR{n}$ are the parameters of the policy and $\pi$ is the probability distribution of sampling action $\bm{a}_{t}$ in state $\bm{s}_{t}$ at time $t$. 
	    Performing action $\bm{a}_{t}$ changes the world state to $\bm{s}_{t+1}$ and the agent receives a reward $r_{t+1}$, associated with the transition $T(\bm{s}_{t},\bm{a}_{t},\bm{s}_{t+1})$. The agent's objective is to maximize the expected return of the policy~\cite{sutton2018reinforcement}, i.e., 
	    \begin{equation}
                \max_{\bm{\theta}}~\mathbb{E}_{\pi_{\bm{\theta}}}\left[R(\bm{s},\bm{a})\right] = 
	        \max_{\bm{\theta}}~\mathbb{E}_{\pi_{\bm{\theta}}} \left[\sum_t r(\bm{s}_{t},\bm{a}_{t})\right].
	    \end{equation}

        In this paper, we have used different  \ac{rl} algorithms and a \ac{bbo} algorithm for policy improvement to show the versatility of our proposed approach. The used algorithms are briefly reviewed as follows.
    
	    \subsubsection{\ac{power}}
            \ac{power}~\cite{Kober2011ML} is an \ac{rl} policy search algorithm inspired by expectation maximization in supervised learning algorithms. It is designed for finite horizons with episodic restarts and uses an average return as a weight instead of a gradient.
    
	    \subsubsection{\ac{sac}}
	        \ac{sac}~\cite{haarnoja2018soft} is an instance of entropy-regularized deep \ac{rl}, which aims to maximize the policy's return while also maximizing entropy. An entropy coefficient is used to control the importance of entropy and is adjusted during training.
	
	    \subsubsection{\ac{ppo}}
	        \ac{ppo}~\cite{schulman2017proximal} is a deep \ac{rl} policy gradient optimization algorithm that clips policy gradient updates to a narrow interval, ensuring the new policy is not too far from the existing one.

	    \subsubsection{\ac{cmaes}}
	        \ac{cmaes}~\cite{hansen2006cma} is a derivative-free method for non-linear or non-convex \ac{bbo} problems in the continuous domain. Instead of using gradient information, \ac{cmaes} makes use of evolutionary computation and an evolution strategy to solve the optimization problem.
	
	\subsection{Riemannian manifold}
		A Riemannian manifold $\mathcal{M}$ is an $n$-dimensional smooth differentiable topological space equipped with a Riemannian metric that locally resembles the Euclidean space $\manR{n}$. The locally Euclidean tangent space $\mathcal{T}_{\bm{X}}\mathcal{M}$ can be constructed around any point $\bm{X} \in \mathcal{M}$. The Riemannian metric,  defined as the positive definite inner product, can be used to generalize the notion of the straight line between two points in Euclidean space by defining the shortest curve between two points in a manifold, which is denoted as a geodesic.
    
        In order to go back and forth between a manifold $\mathcal{M}$ and a tangent space $\mathcal{T}_{\bm{X}}\mathcal{M}$, we require two distance-preserving mapping functions (operators). These operators are (\emph{i}) the exponential map $\mathrm{Exp}_{\bm{X}}: \mathcal{T}_{\bm{X}}\mathcal{M} \rightarrow \mathcal{M}$, and its inverse (\emph{ii}) the logarithmic map $\mathrm{Log}_{\bm{X}}: \mathcal{M} \rightarrow \mathcal{T}_{\bm{X}}\mathcal{M}$ as depicted in Fig.~\ref{fig:both}~(a). It is possible to show that exponential and logarithmic maps are (locally) bijective~\cite{jost2008riemannian}, which makes it possible to do the calculations about the non-Euclidean manifold space on the tangent space and project back the results. 
        Another essential concept in differential geometry is parallel transport $\Gamma_{\bm{X} \rightarrow \bm{Y}}$, allowing for calculations and comparisons of vectors located on different tangent spaces to be carried out by moving vectors through a connecting geodesic. This method preserves the inner product between transported vectors.

         A Gaussian distribution on Riemannian manifolds is defined as in~\cite{calinon2020gaussians}
         \begin{equation}
             \mathcal{N}_{\mathcal{M}}(\bm{Q}|\bm{X},\bm{\Sigma}) = \big((2\pi)^d |\bm{\Sigma}| \big)^{-\frac{1}{2}}e^{\mathrm{Log}_{\bm{X}}(\bm{Q})\bm{\Sigma}^{-1}{\mathrm{Log}_{\bm{X}}(\bm{Q})}},
         \end{equation}

        where $\bm{X} \in \mathcal{M}$ , the covariance $\bm\Sigma$ is defined on $\mathcal{T}_{\bm{X}}\mathcal{M}$ and $\bm{Q}\in\mathcal{M}$. For more details about Gaussian distributions on manifolds in the context of robotics, we refer the interested reader to~\cite{calinon2020gaussians}.
\section{Policy Parameterization on Tangent Space}  
    Recently, the topic of learning using Riemannian geometry tools has become the focus of researchers in the field of robot learning~\cite{bronstein2017geometric,  abu2021probabilistic, calinon2020gaussians, abu2020geometry}. An example of this is when considering a robot's operational space, where its end-effector pose consists of a Cartesian position (Euclidean part) and orientation (non-Euclidean part). It is common to apply learning in this space since it allows for kinematic redundancy and the ability to transfer a learned policy from one robot to another robot with different anatomy~\cite{zeestraten2018programming}.
    
\begin{figure}[!t]
    \centering
	\def\svgwidth{\linewidth}
	{\fontsize{8}{8}
		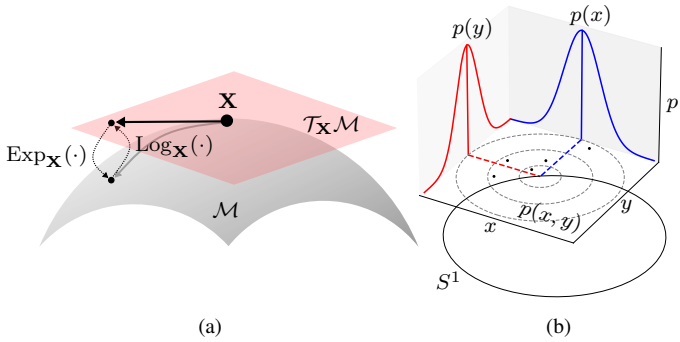}
	\caption{(a) The gray surface represents a manifold $\mathcal{M}$, and the red plane represents the tangent space $\mathcal{T}_{\bm{X}}\mathcal{M}$. The exponential/logarithmic mapping tools between the two spaces are shown. (b) Sampling an $S^1$ manifold from a Gaussian distribution, where the mean is on $S^1$, but the drawn samples may not be, like the points in $p(x,y)$.}
	\label{fig:both}
    \end{figure}

    Gaussian policy parameterization has a limitation when it comes to representing non-Euclidean data like orientation, stiffness, or manipulability, as the distribution parameters (both mean and covariance) do not always obey the nature of the manifold’s curvature space.
    The problem with sampling non-Euclidean from a Gaussian distribution is illustrated in Fig.~\ref{fig:both}~(b) for the $S^1$ manifold, i.e., the circumference of the unit circle. Picking a point on the manifold to be the mean of the normal distribution, samples can still be drawn from outside the manifold, as illustrated in the figure. Normalization of the sample can map it back to the unit circle manifold at the cost of accuracy. The same argument is applicable to other manifolds like $\manS{3}$ embedded in $\manR{4}$. To this point, using Gaussian policy parameterization like \ac{sac}~\cite{haarnoja2018soft} or \ac{ppo}~\cite{schulman2017proximal} on non-Euclidean manifold data like quaternions will require normalizing the predicted profiles. This kind of post-processing is an approximation that could affect learning accuracy.

	Having a framework allowing for well-established and stable learning algorithms on Euclidean space to be transferred to other geometrical spaces with relative ease and reasonable computational costs is beneficial. This enables part of the achievements and progress that have been made on Euclidean space to be directly applicable to non-Euclidean spaces.
    \ac{grl} is based on applying parameterization on a constant tangent space, where there is no need to parallel transport the policy being learned from one tangent space to another. Doing so is not trivial for some parameterization schemes. At the same time, we must obey the formulation of the Riemannian geometry, which is locally bijective.
    
    Thus, let us consider $\mathcal{M}$ and $\mathcal{N}$ as two Riemannian manifolds, where $\bm{M}_P, \bm{M}_L \in \mathcal{M}$ and $\bm{N}_P, \bm{N}_L \in \mathcal{N}$. Conceptually, when \ac{grl} is used to learn data that correspond to a Riemannian manifold $\mathcal{M}$, we utilize the tangent space in two ways: a constant tangent space $\mathcal{T}_{\bm{M}_{P}}\mathcal{M}$ for parameterization, and a local tangent space $\mathcal{T}_{\bm{M}_{L}}\mathcal{M}$ for mapping to manifold actions. The manifolds' data which are indexed with $P$ represent the data points where the parameterization tangent spaces are established, and the ones indexed with $L$ represent the data points where the moving local tangent spaces are established.

    In the case where consecutive actions $i$ and $i+1$ are local to one another (such as learning a smooth trajectory of orientations), the local tangent space is situated on the previous action (e.g., predicting the orientation at time $i+1$ means situating the local tangent space on the predicted orientation at time $i$). Parallel transport must then be employed to move the parameterized vectors to the local tangent spaces. But in the case where the rollout consists of a single action and the different rollouts are independent of each other (\eg Wahba problem), we locate both the parameterization tangent space and local tangent space onto the same point. Note that the parameterization tangent space is never moved itself; the policy is learned on a fixed, constant tangent space. In either case, we map the result back to the manifold once the vector is moved to the local tangent space.

    In the general setting of learning a manipulation task, it is common to have state and action data from different manifolds, in other words, having a composite manifold, which is defined as the Cartesian product of the manifolds. For example, the state $\bm{s} \in \mathcal{M} \times \mathcal{N}$ and the action ${\bm{a} \in \mathcal{M} \times \mathcal{N}}$.

    The parameterization tangent space of the composite manifold is represented as $\mathcal{P}$: $\mathcal{T}_{(\bm{M}_{P},\bm{N}_{P})}(\mathcal{M}\times \mathcal{N})$, while the local tangent space of the composite manifold is represented by $\mathcal{L}$: $\mathcal{T}_{(\bm{M}_{L},\bm{N}_{L})}(\mathcal{M}\times \mathcal{N})$.

    The state at time $t$ as a composite manifold state is represented as follows: 

    \begin{equation}
        \bm{s}_t = (\bm{S}_{\mathcal{M},t}, \bm{S}_{\mathcal{N},t}),
        \label{eq:state_comp}
    \end{equation}
    where $\bm{S}_{\mathcal{M},t}$ and $\bm{S}_{\mathcal{N},t}$ are the state parts that belong to each of the two manifolds $\mathcal{M}$ and $\mathcal{N}$, respectively. The action $\boldsymbol{\mathfrak{a}}_{P,t}$ on the composite parameterization tangent space at time $t$ is represented as follows:
    \begin{equation}
        \boldsymbol{\mathfrak{a}}_{P,t} = [\boldsymbol{\mathfrak{a}}_{P_{\mathcal{M}},t} \| \boldsymbol{\mathfrak{a}}_{P_{\mathcal{N}},t}], 
     \label{eq:action_comp_tangent_P}
    \end{equation}    
    while the action $\boldsymbol{\mathfrak{a}}_{L,t}$ on the composite local tangent space is given by
    \begin{equation}
        \boldsymbol{\mathfrak{a}}_{L,t} = [\boldsymbol{\mathfrak{a}}_{L_{\mathcal{M}},t} \| \boldsymbol{\mathfrak{a}}_{L_{\mathcal{N}},t}], 
     \label{eq:action_comp_tangent_L}
    \end{equation}
where the subscripts ${}_{_{\mathcal{M}},t}$ and ${}_{_{\mathcal{N}},t}$ denote the part of the action coming from manifolds $\mathcal{M}$ and $\mathcal{N}$, respectively.
The $[\cdot\|\cdot]$ is a concatenation operator. Intuitively, the prediction on the tangent space allows us to ``stack'' different manifolds into a unique vector. Afterward, we use the parallel transport operator to transport the action vector from $\mathcal{P}$ to $\mathcal{L}$ at $t$ as
    \begin{equation}
       \boldsymbol{\mathfrak{a}}_{L,t} = \Gamma_{\mathcal{P} \rightarrow \mathcal{L}}(\boldsymbol{\mathfrak{a}}_{P,t}).
        \label{eq:action_parallel_transport}
    \end{equation}
Subsequently, we project this local action vector to the composite manifold as follows
    \begin{equation}
    \begin{split}
        \bm{a}_{t} &\begin{aligned}[t]&= (\bm{A}_{\mathcal{M},t} , \bm{A}_{\mathcal{N},t}) \\&= (\mathrm{Exp}_{\bm{M}_{L,t}}(\boldsymbol{\mathfrak{a}}_{L_{\mathcal{M}},t}) , \mathrm{Exp}_{\bm{N}_{L,t}}(\boldsymbol{\mathfrak{a}}_{L_{\mathcal{N}},t})).\end{aligned}
    \end{split}
        \label{eq:action_comp_manifold}
    \end{equation}
    
    The policy $\pi_{\bm{\theta}}$ predicts the action on the parameterization tangent space $\boldsymbol{\mathfrak{a}}_{P}$ according to the current state $\bm{s}$ as  follows:
	\begin{equation}
    \begin{split}
        \pi_{\bm{\theta}} & (\boldsymbol{\mathfrak{a}}_{P}| \bm{s}) = \left[\pi_{\bm{\theta}_\mathcal{M}}\left(\boldsymbol{\mathfrak{a}}_{P_{\mathcal{M}}}|\bm{S}_{\mathcal{M}}\right) \| \pi_{\bm{\theta}_\mathcal{N}}(\boldsymbol{\mathfrak{a}}_{P_{\mathcal{N}}}|\bm{S}_{\mathcal{N}})\right]
     \label{eq:policy}
    \end{split}
    \end{equation}
    where $\bm{\theta} = [\bm{\theta}_\mathcal{M} \| \bm{\theta}_\mathcal{N}]$ is the concatenation of parameters for the manifolds, respectively. At each time $t$, an action $\boldsymbol{\mathfrak{a}}_{P,t} = [\boldsymbol{\mathfrak{a}}_{P_{\mathcal{M}},t} \| \boldsymbol{\mathfrak{a}}_{P_{\mathcal{N}},t}]$ is drawn from the policy~\eqref{eq:policy} as

    \begin{equation}
        \boldsymbol{\mathfrak{a}}_{P,t} \sim \left[\pi_{\bm{\theta}_\mathcal{M}}\left(\boldsymbol{{\mathfrak{a}}}_{P_{\mathcal{M}},t}|\bm{S}_{\mathcal{M},t}\right) \| \pi_{\bm{\theta}_\mathcal{N}}(\boldsymbol{\mathfrak{a}}_{P_{\mathcal{N}},t}|\bm{S}_{\mathcal{N},t})\right].\label{eq:pp1}
    \end{equation}
    
    The action $\boldsymbol{\mathfrak{a}}_{P,t}$ is converted in a manifold action $\bm{a}_{t}$ using~\eqref{eq:action_comp_manifold}, and the agent performs the resulting manifold action on the environment. This causes the state to transition from $\bm{s}_{t}$ to $\bm{s}_{t+1}$. The expected return captures the expected quality of the policy
	\begin{equation}
    \begin{split}
	    \mathbb{E}_{\pi_{\bm{\theta}}}&\left[\sum_t r(\bm{s}_t,\bm{a}_t)\right] = \\
        & \mathbb{E}_{\pi_{\bm{\theta}}}\left[\sum_t r((\bm{S}_{\mathcal{M},t},\bm{S}_{\mathcal{N},t}),(\bm{A}_{\mathcal{M},t},\bm{A}_{\mathcal{N},t}))\right]
	    \label{eq:reward}.
     \end{split}
     \end{equation}
	
	\begin{algorithm}[t!]
		\textbf{Input:} initial state $\bm{s}_0$, initial parameters $\bm{\theta}$, $\bm{M}_{P},\bm{M}_{0} \in \mathcal{M}$, $\bm{N}_{P},\bm{N}_{0} \in \mathcal{N}$, where $\bm{M}_P$, and $\bm{N}_P$ are the centers of the composite parameterization tangent space. $\bm{M}_0$, $\bm{N}_0$ are the centers of the initial composite local tangent space and the \ac{rl} algorithm $\alpha$.
		\begin{algorithmic}[1]
			\While{\texttt{!stop\_condition(}$\alpha$\texttt{)}}						
		          \State{ $\pi_{\bm{\theta}}(\boldsymbol{\mathfrak{a}}_{P}| \bm{s}) \leftarrow$ \texttt{get\_policy(}$\bm{\theta}$, $\alpha$\texttt{)}\Comment{eq. \eqref{eq:policy}}}
				\State{$R(\bm{s},\bm{a})\leftarrow  0$ \Comment{cumulative reward}}
				\For{$t = 0,\ldots, T-1$ } 
					\State{$\bm{s}_{t} \leftarrow (\bm{S}_{\mathcal{M},t},\bm{S}_{\mathcal{N},t})$ \Comment{state composition \eqref{eq:state_comp}}}
					\State \multiline{%
						$\boldsymbol{\mathfrak{a}}_{\mathcal{P},t} \sim \pi_{\bm{\theta}}(\boldsymbol{\mathfrak{a}}_{\mathcal{P},t}|\bm{s}_{t})$ \Comment{tangent space action \eqref{eq:pp1}}}
					\State{$\boldsymbol{\mathfrak{a}}_{\mathcal{P},t} = \left[ \boldsymbol{\mathfrak{a}}_{P_{\mathcal{M}},t} \|  \boldsymbol{\mathfrak{a}}_{P_{\mathcal{N}},t} \right]$ \Comment{act. concatenation \eqref{eq:action_comp_tangent_P}}} 
                        \State{$\boldsymbol{\mathfrak{a}}_{\mathcal{L},t} \leftarrow \Gamma_{\mathcal{P} \rightarrow \mathcal{L}_{t}}(\boldsymbol{\mathfrak{a}}_{\mathcal{P},t})$\Comment{act. par. trans. \eqref{eq:action_parallel_transport}}}
                    \State{$\boldsymbol{\mathfrak{a}}_{\mathcal{L},t} = \left[ \boldsymbol{\mathfrak{a}}_{L_{\mathcal{M}},t} \|  \boldsymbol{\mathfrak{a}}_{L_{\mathcal{N}},t} \right]$ \Comment{act. concatenation \eqref{eq:action_comp_tangent_L}}}
				    
                    \State{$\bm{a}_{t} \leftarrow (\mathrm{Exp}_{\bm{M}_{L,t}}(\boldsymbol{\mathfrak{a}}_{L_{\mathcal{M}},t}),\mathrm{Exp}_{\bm{N}_{L,t}}(\boldsymbol{\mathfrak{a}}_{L_{\mathcal{N}},t}))$ \Comment{manifold act. \eqref{eq:action_comp_manifold}}}
                    
                    \State{ $(\bm{M}_{t+1},\bm{N}_{t+1}) \leftarrow (\bm{S}_{\mathcal{M},t},\bm{S}_{\mathcal{N},t})$}
                    
				    \State $\bm{s}_{t+1} \leftarrow$ \texttt{execute\_on\_robot(}$\bm{a}_{t}$\texttt{)}
						
					\State \multiline{%
						$R(\bm{s},\bm{a}) \leftarrow R(\bm{s},\bm{a})  + r(\bm{s}_t,\bm{a}_t)$ }
				\EndFor
				\State $\bm{\theta} \leftarrow$ \texttt{improve\_policy(}$\bm{\theta}$, $R(\bm{s},\bm{a})$, $\alpha$\texttt{)}
			\EndWhile
		\end{algorithmic}
		\caption{\acf{grl}}
	\end{algorithm}
	As shown in Algorithm 1, the initial state, the centers of the composite of two tangent spaces, and the \ac{rl} algorithm are used as input. In line 2, the \ac{rl} algorithm generates a policy structure $\pi_{\bm{\theta}}(\boldsymbol{\mathfrak{a}}\text{$|$} \bm{s})$ with the current parameters $\bm{\theta}$. This policy operates in the composite parameterization tangent space established at the composite point $(\bm{M}_{P},\bm{N}_{P})$ given as input to the \ac{rl} algorithm. The parameterization is based on the composite current state in line 5, as it is passed to the policy in line 6 to sample the composite action $\boldsymbol{\mathfrak{a}}_P$ on the parameterization tangent space. This action (defined in line 7) is parallel transported to the current local composite tangent space (line 8) and gives the action vector defined in line 9. Line 10 maps the composite tangent space action into the composite manifold. After that, the local composite tangent space is updated to the current composite state (line 11), the action is executed by the agent, and the state is updated (line 12). At each step in the rollout, the total reward is updated by accumulating the immediate rewards (line 13). After one rollout is finished, the quality of the policy is measured using the rollout total reward, which is passed to the \ac{rl} algorithm to proceed with learning (line 15). This procedure is repeated until the stopping criteria, depending on the used \ac{rl} algorithm, is met (line 1).

 \subsection{{Learning on the $\manS{3}$ manifold}}
 Orientations are commonly represented using rotation matrices, Euler angles, or unit quaternions. Euler angles are a minimal orientation representation (requiring only three parameters) but suffer from the singularity problem~\cite{murray2017mathematical}. Unit quaternions hold an advantage over rotation matrices due to requiring fewer parameters (4 instead of 9) and are therefore commonly used to represent rotation in robotic applications. The unit quaternion representation belongs to the 3-sphere manifold, denoted as $\manS{3}$~\cite{murray2017mathematical}. Therefore, applying current reinforcement learning algorithms designed for Euclidean space to learn an orientation policy is not straightforward as it normally involves approximations to account for the underlying manifold structure.
	
    In this section, we focus on orientation learning represented by unit quaternions. A quaternion, denoted as $\bm{Q}$, is a tuple $(v,\mathbf{u})$ composed of a scalar $v$ and a three-dimensional vector $\mathbf{u}=(x,y,z)$. Unit quaternions have a norm of one and belong to $\manS{3}$. The hypersphere $\manS{3}$ has a double-covering of $\manSO{3}$, meaning that for every rotation in $\manSO{3}$ there exist two quaternions that can represent it ($\bm{Q}$ and $-\bm{Q}$). In this section, actions can be represented as unit quaternions, and the learning is carried out on a single hemisphere; in case we have a prediction $\bm{Q}$ on the other hemisphere, we flip the prediction by using $-\bm{Q}$.
     
     In order to effectively utilize Gaussian distribution calculations for unit quaternions, it is necessary to take into account their geometric properties. The objective is to maximize the expected reward as defined in equation \eqref{eq:reward}. 
	    
	    In this context, we define $\mathcal{M}\equiv \manS{3}$, and consider unit quaternions $\bm{Q}=(v, \bm{u})$, $\bm{Q}_1=(v_1, \bm{u}_1)$, ${\bm{Q}_2=(v_2, \bm{u}_2)\in \manS{3}}$, and $\boldsymbol{\mathfrak{q}},\boldsymbol{\mathfrak{u}}\in\mathcal{T}_{\bm{Q}}\manS{3}$ The logarithmic map, denoted as $\mathrm{Log}_{\bm{Q}_{1}}(\cdot)$ is redefined to map $\bm{Q}_2$ into $\mathcal{T}_{\bm{Q}_{1}}\manS{3}$ \eg ${\mathrm{Log}_{\bm{Q}_{1}}(\cdot): \manS{3} \mapsto \mathcal{R}^4}$~\cite{absil2007optimization} as 
    \begin{equation}
        \mathrm{Log}_{\bm{Q}_1}(\bm{Q}_{2})= \frac{\bm{Q}_{2}-({\bm{Q}_1}\trsp\bm{Q}_{2}){\bm{Q}_1}}{\|\bm{Q}_{2}-({\bm{Q}_1}\trsp\bm{Q}_{2}){\bm{Q}_1}\|}d({\bm{Q}_1},\bm{Q}_{2}),
        \label{eq:log}
    \end{equation}

    where $\|\cdot\|$ defines the norm of a vector, and the distance between two unit quaternions is defined as follows
    \begin{equation}
        d(\bm{Q}_1, \bm{Q}_2)= \arccos(\bm{Q}_1\trsp \bm{Q}_2),
        \label{eq:dist}
    \end{equation}
    
    For example, if the reward function is $\exp(-d)$, where $d$ is the distance between two unit quaternions, then equation \eqref{eq:dist} is used to calculate the distance on the tangent space.
    
	In \eqref{eq:action_comp_manifold}, the exponential map, denoted as $\mathrm{Exp}_{\bm{Q}_1}(\cdot)$, is redefined to project actions from the current local tangent space into the hypersphere $\manS{3}$, \eg ${\mathrm{Exp}_{\bm{Q}_1}(\cdot):\mathcal{R}^4 \mapsto \manS{3}}$~\cite{absil2007optimization}

    \begin{equation}
    \mathrm{Exp}_{\bm{Q}_1}(\boldsymbol{\mathfrak{q}})= {\bm{Q}_1}\cos(\|\boldsymbol{\mathfrak{q}}\|)+\frac{\boldsymbol{\mathfrak{q}}}{\|\boldsymbol{\mathfrak{q}}\|}\sin(\|\boldsymbol{\mathfrak{q}}\|).
        \label{eq:exp}
    \end{equation}
    
     Parallel transport in~\eqref{eq:action_parallel_transport} is redefined as in~\cite{absil2007optimization} :
    \begin{equation}
    \begin{split}
        \Gamma_{{\bm{Q}_1} \rightarrow {\bm{Q}_2}}(\boldsymbol{\mathfrak{q}}) = &(-{\bm{Q}_1}\sin(\|\boldsymbol{\mathfrak{u}}\|)\overline{\boldsymbol{\mathfrak{u}}}\trsp + \overline{\boldsymbol{\mathfrak{u}}}\cos(\|\mathfrak{u}\|)\overline{\boldsymbol{\mathfrak{u}}}\trsp \\
        &+ (\bm{I}-\overline{\boldsymbol{\mathfrak{u}}}~\overline{\boldsymbol{\mathfrak{u}}}\trsp)) \boldsymbol{\mathfrak{q}}
    \end{split}
    \end{equation}
    with $\overline{\boldsymbol{\mathfrak{u}}} = \frac{\boldsymbol{\mathfrak{u}}}{\|\boldsymbol{\mathfrak{u}}\|}$, and $\boldsymbol{\mathfrak{u}} =\mathrm{Log}_{\bm{Q}_1}({\bm{Q}_2})$.

\subsection{Learning on the $\manSPD{\MakeLowercase{d}}$ manifold} 
Data such as stiffness, manipulability, and covariance ellipsoids/matrices play a vital role in robotic manipulation. Such data belong to the space of \ac{spd} matrices. However, effectively learning these data using \ac{rl} algorithms is challenging due to the need for approximations to conform to the manifold geometry. Typically, Cholesky decomposition is employed to guarantee that the predicted matrix remains \ac{spd} \cite{abu2018force}.

A matrix $\bm{\Sigma}$ belongs to the space $\manSPD{d}$ if it satisfies two conditions:  symmetry (\ie $\bm{\Sigma}=\bm{\Sigma}\trsp$) and positive definiteness $\bm{v}\trsp \bm{\Sigma} \bm{v}>0,\, \forall\,\, \text{nonzero vectors}\,\, \bm{v}$. To express manifold operators for \ac{spd} matrices as outlined in~\cite{pennec2006riemannian,sra2015conic}, we introduce the notation $\bm{\Sigma}_1$, $\bm{\Sigma}_2$, $\bm{W} \in \manSPD{d}$ and $\boldsymbol{\mathfrak{w}} \in \mathcal{T}_{\bm{\Sigma}}\manSPD{d}$.
The exponential map, denoted as $\mathrm{Exp}_{\bm{\Sigma}}(\cdot)$ in \eqref{eq:action_comp_manifold}, is redefined to project actions from the current local tangent space to the $\manSPD{d}$ manifold
\begin{equation}
    \text{Exp}_{\bm{\Sigma}} (\boldsymbol{\mathfrak{w}}) = \bm{\Sigma}^\frac{1}{2} \text{expm} \left( \bm{\Sigma}^{-\frac{1}{2}} \boldsymbol{\mathfrak{w}} \bm{\Sigma}^{-\frac{1}{2}} \right) \bm{\Sigma}^{\frac{1}{2}}
\end{equation}
The logarithmic map, denoted as $\mathrm{Log}_{\bm{\Sigma}}(\cdot)$, is redefined to map $\bm{W}$ to $\mathcal{T}_{\bm{\Sigma}}\manSPD{d}$
\begin{equation}
    \text{Log}_{\bm{\Sigma}}(\bm{W}) = \bm{\Sigma}^\frac{1}{2} \text{logm} \left( \bm{\Sigma}^{-\frac{1}{2}} \bm{W} \bm{\Sigma}^{-\frac{1}{2}} \right) \bm{\Sigma}^{\frac{1}{2}}
\end{equation}
Parallel transport is defined as
\begin{equation}
    \mathcal{T}_{\bm{\Sigma}_1 \to \bm{\Sigma}_2}(\boldsymbol{\mathfrak{w}}) = \bm{\Sigma}_2^{\frac{1}{2}} \bm{\Sigma}_1^{-\frac{1}{2}} \boldsymbol{\mathfrak{w}} ~ \bm{\Sigma}_1^{-\frac{1}{2}}\trsp \bm{\Sigma}_2^{\frac{1}{2}}\trsp
\end{equation}
The distance between two \ac{spd} matrices is defined as follows
\begin{equation}
    d(\bm{\Sigma},\bm{W}) = \left|\left| \text{logm} \left( \bm{\Sigma}^{-\frac{1}{2}} \bm{W} \bm{\Sigma}^{-\frac{1}{2}} \right) \right|\right|_F
    \label{eq:16}
\end{equation}
where $||\cdot||_F$ is the Frobenius norm.

An important feature of $\manSPD{d}$ is that it has no cut locus, resulting in a bijective mapping over the entire manifold space~\cite{pennec2006riemannian}.

When parameterizing \ac{spd} matrices, two approaches were used: vectorization via both Cholesky factorization and Mandel notation. In the Cholesky factorization approach, an \ac{spd} matrix $\Sigma$ is represented as the product of its Cholesky factor $\bm{L}$ and its transpose, \ie $\Sigma = \bm{L}^\top \bm{L}$. The vectorization is then performed on the upper triangle elements of $\bm{L}$ for learning purposes. Alternatively, in the Mandel notation approach, an \ac{spd} matrix $\Sigma$ can be defined using a specific vector representation. For example, in the case of $3\times3$ \ac{spd} matrix $[\Sigma] = [\Sigma_{11}, \Sigma_{22}, \Sigma_{33}, \sqrt{2}\Sigma_{23}, \sqrt{2}\Sigma_{13}, \sqrt{2}\Sigma_{12}]^\top$. In \ac{grl} implementation for \ac{spd} data, we utilized the Mandel notation to reduce the dimensionality of the data. Additionally, we used Mandel notation as a baseline, referred to as "Mandel," where we find the nearest \ac{spd} matrix to the predicted symmetric matrix. We experimentally evaluate both vectorization approaches in Section~\ref{subsec:spd_wahba} and Section~\ref{subsec:spd_traj_learn}.

\section{Experimental results}\label{sec:results}
	
	Experiments have been carried out in simulated environments (Wahba~\cite{wahba1965least} and trajectory learning problems), as well as a real setup involving a physical robot performing the Ball-in-a-hole task. Several \ac{rl} and policy improvement algorithms have been tested: 
	\begin{itemize}
	    \item deep \ac{rl} algorithms like \ac{sac}~\cite{haarnoja2018soft} and \ac{ppo}~\cite{schulman2017proximal},
	    \item the expectation-maximization inspired \ac{power} algorithm~\cite{kober2011policy}, and
	    \item the \ac{bbo}-based \ac{cmaes} algorithm~\cite{hansen2006cma}.
	\end{itemize}
	Our research question is about the gains of considering the geometry of non-Euclidean data (\eg orientation, stiffness, or manipulability) in \ac{rl} algorithms based on Gaussian distributions and how they compare with the common approximation solutions (\eg normalization and Cholesky decomposition) or solutions based on other distributions like Bingham.
	\subsection{simulation experiments}\label{sec:simulations}
	\subsubsection{Quaternion Wahba problem}\label{subsec:wahba}
	
	\begin{figure}[!t]
        \centering
	    \def\svgwidth{\linewidth}
	    {\fontsize{8}{8}
		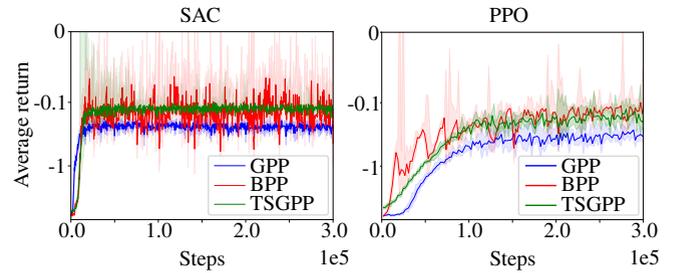}
    	\caption{Quaternion Wahba domain results for \ac{sac} and \ac{ppo} using \ac{gpp}, \ac{grl}, and \ac{bpp}. The mean (solid lines) and the standard deviation (shaded regions) are calculated over five different seeds.}
    	\label{fig:wahba100_sp}
    \end{figure}
	
	The Wahba problem, first proposed by Grace Wahba in 1965~\cite{wahba1965least}, is about finding the best rotation between two Euclidean coordinate systems that aligns two sets of noisy 3-dimensional vector observations. The original motivation for this problem was to estimate satellite altitudes using vectors from different frames of reference, but it was later applied to other research fields as well.

\begin{figure}[!t]
        \centering
	    \def\svgwidth{\linewidth}
	    {\fontsize{8}{8}
		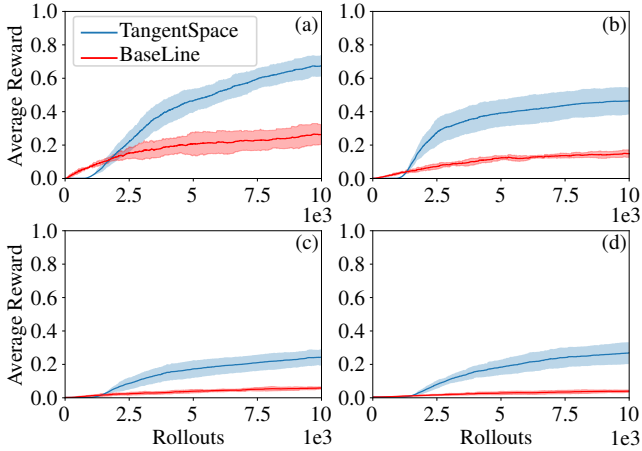}
	    \caption{Four instances of the variation of the Wahba problem with different sizes (complexity) solved by the \ac{power} algorithm. The size for each is (a) 10, (b) 12, (c) 14, and (d) 16. The mean (solid lines) and the standard deviation (shaded regions) are calculated over five different seeds.}
    	\label{fig:PoWER_wahba}
    \end{figure}

\begin{figure}[!t]
        \centering
	    \def\svgwidth{\linewidth}
	    {\fontsize{8}{8}
		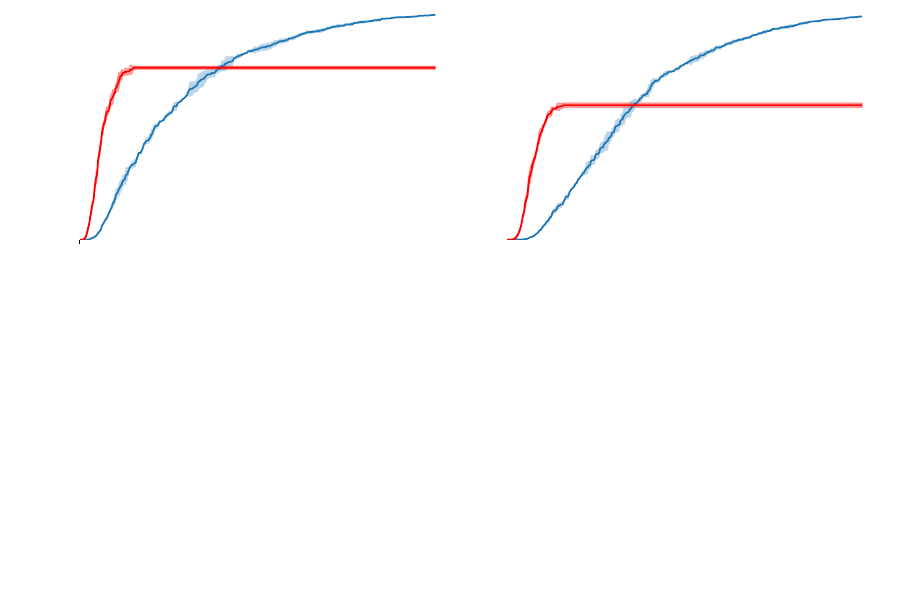}
	    \caption{Four instances of the variation of the Wahba problem with different sizes (complexity) solved by the \ac{cmaes} algorithm. The size for each is (a) 10, (b) 12, (c) 14, and (d) 16. The mean (solid lines) and the standard deviation (shaded regions) are calculated over five different seeds.}
    	\label{fig:CMAES_Wahba}
    \end{figure}

    The cost function defines attempts to minimize the difference between sets of vectors  ($\mathbf{y}_i \in Y, \mathbf{z}_i \in Z$) by finding a rotation $\mathbf{R} \in \manSO{3}$
    \begin{equation}
        J(\mathbf{R})=\frac{1}{2}\sum_{k=1}^N a_k \|\mathbf{z}_k - \mathbf{R}\mathbf{y}_k\|^2.
    \end{equation}
    where $a_k$ are the weights for each observation. In our case, orientation is represented by unit quaternions. Our experiments use a set of random 3-dimensional vectors and their corresponding rotated ones as the state. The predicted unit quaternion $\bm{\hat{Q}}$ is compared to the original rotation $\bm{Q}$ with a reward given by $r = -{d(\bm{Q},\bm{\hat{Q}}})$ as in Fig.~\ref{fig:wahba100_sp}, or by $r = e^{-{d(\bm{Q},\bm{\hat{Q}}})}$ as in Fig.~\ref{fig:PoWER_wahba} and Fig.~\ref{fig:CMAES_Wahba}, where $d(\bm{Q}_t,\bm{\hat{Q}}_t)$ is the distance between two unit quaternions as given by equation \eqref{eq:dist}.

	Figure \ref{fig:wahba100_sp} shows the results of learning the orientation represented as a unit quaternion using \acf{gpp}, \acf{grl}, and \acf{bpp}~\cite{james2022bingham}. The quality of the learned policy using \ac{grl} was better than \ac{gpp} for both \ac{sac}~\cite{haarnoja2018soft} and \ac{ppo}~\cite{schulman2017proximal}, while compared to \ac{bpp} a slightly better policy was learned for \ac{sac} and a comparable policy was learned for \ac{ppo}.
	
	We also used a less complex variation of the Wahba problem by limiting the number of learning orientations to 10, 12, 14, and 16 for \ac{power} and \ac{cmaes}. As shown in Fig. \ref{fig:PoWER_wahba} and Fig. \ref{fig:CMAES_Wahba}, our goal from these experiments is to show the importance of avoiding approximation (normalization) when learning unit quaternions. The results of \ac{grl} are significantly better than the \ac{gpp} results. 

    \begin{figure}[!t]
        \centering
        \def\svgwidth{\linewidth}
        {\fontsize{8}{8}
            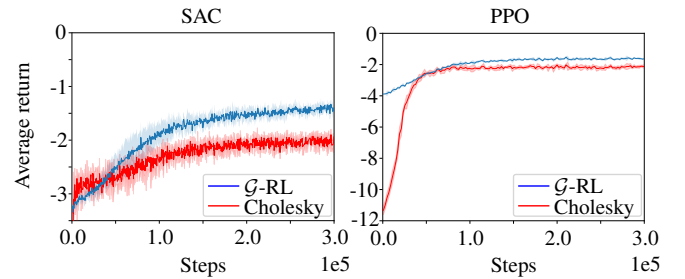}
        \caption{Illustrates the quality of learning a policy, utilizing \ac{sac} on the left and \ac{ppo} on the right, in solving a variation of the Wahba problem. The policy's objective is to predict \ac{spd} matrices that represent the stiffness coefficient of spring, as well as a set of vectors representing positional displacements. By manipulating the vector set with the \ac{spd} matrices, the spring force is determined for each displacement. The resulting curve of our \ac{grl} approach is in blue, while the resulting baseline curve, using Cholesky decomposition, is in red. The solid lines indicate the mean performance across five different random seeds, while the shaded regions represent the standard deviation.}
        \label{fig:wahba100_spd}
    \end{figure}
        
        \subsubsection{\ac{spd} Matrix Wahba Problem}\label{subsec:spd_wahba}
        In addition to quaternions, the Wahba problem was also implemented with \ac{spd} matrices manipulating a set of random 3-dimensional vectors. One can think of this problem as a spring system, where the \ac{spd} represents the stiffness coefficient of spring and the vector set represent positional displacements. Manipulating the vector set with the \ac{spd} provides the spring force at each displacement.
        
        The reward for this version of the problem is given by
        $r=-d(\bm{W},\widehat{\bm{W}})$,
        where
        $d(\bm{W},\widehat{\bm{W}})$
        is the affine invariant distance between the original \ac{spd} matrix $\bm{W}$ and the predicted \ac{spd} matrix $\widehat{\bm{W}}$ given by equation~\eqref{eq:16}.
        
         Fig. \ref{fig:wahba100_spd} shows the progression in the quality of learning a policy (specifically, a variation of the Wahba problem) over time. This is depicted using \hbox{\ac{sac}} on the left and \hbox{\ac{ppo}} on the right. The policy in question predicts the \hbox{\ac{spd}} that represents the spring's stiffness coefficient, which operates on a vector set that represents positional displacements. The force of the spring at each displacement is derived when the vector set is manipulated with the \hbox{\ac{spd}}. In order to ensure a comprehensive comparison, the Cholesky decomposition method, represented in red, is applied as a baseline against our \hbox{\ac{grl}} approach, which is illustrated in blue. Both the mean values (denoted by the solid lines) and the standard deviation (shown via the shaded regions) are computed over five different seeds. As depicted in the figure, the quality of applying \hbox{\ac{rl}} algorithms (\hbox{\ac{sac}} and \hbox{\ac{ppo}}) using \hbox{\ac{grl}} is obviously higher than what is achieved using Cholesky.

\begin{figure}[!t]
            \centering
    	    \def\svgwidth{\linewidth}
    	    {\fontsize{8}{8}
                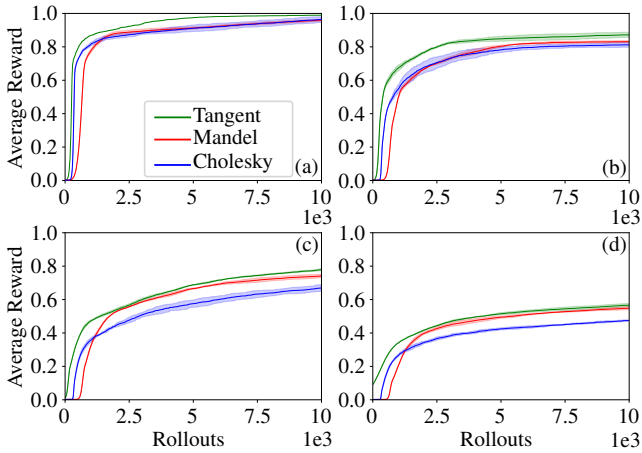}
        	\caption{The \ac{power} algorithm is employed to solve four instances of a variation \ac{spd} Wahba problem, each with a different size indicating varying levels of complexity. The sizes of the instances are (a) 3, (b) 6, (c) 9, and (d) 12. The solid lines represent the mean performance across five different random seeds, while the shaded regions indicate the standard deviation.}
        	\label{fig:PoWER_wahba_spd}
        \end{figure}

\begin{figure}[!t]
            \centering
    	    \def\svgwidth{\linewidth}
    	    {\fontsize{8}{8}
                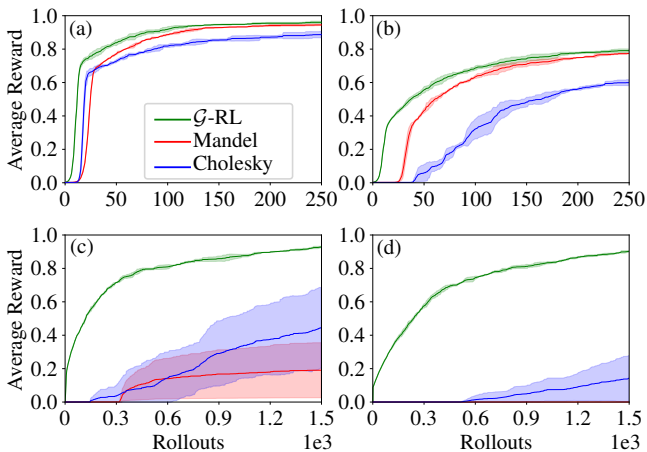}
        	\caption{The \ac{cmaes} algorithm is employed to solve four instances of a modified \ac{spd} Wahba problem, each with a different size indicating varying levels of complexity. The sizes of the instances are as follows: (a) 3, (b) 6, (c) 9, and (d) 12. The solid lines represent the mean performance across five different random seeds, while the shaded regions indicate the standard deviation. Note: the number of rollouts in (c) and (d) is increased to 1500 rollouts to show the significance of the difference between the proposed algorithm and the baseline}
        	\label{fig:CMAES_wahba_spd}
        \end{figure}

        Testing a simpler version of this problem allows for the opportunity to also evaluate alternative parameterization methods for \ac{spd} data, including using Cholesky factorization and Mandel's notation, both of which were evaluated against tangent space parameterization. As seen in Fig. \ref{fig:PoWER_wahba_spd} and Fig. \ref{fig:CMAES_wahba_spd},
        \ac{power} and \ac{cmaes} show that \ac{grl} holds a slight advantage against these other parameterization methods, except for more complex \ac{cmaes} problems where \ac{grl} learns significantly better. Problem sizes 9 and 12 were evaluated under more rollouts to ensure the significance of the comparison. 

	\subsubsection{Orientation Trajectory learning problem}\label{subsec:traj_learn}
    
	Some manipulation learning problems require learning a desired trajectory of the end-effector pose. In this section, we focus on learning a trajectory of orientations where a policy is trained to follow a well-defined trajectory. The current state (orientation) at time $t$ is an input to the policy, and the policy decides what the next (state) orientation at time $t+1$ should be. The reward captures how close the learned trajectory is to the target one, $r = {\sum_{t=1}^{T}}e^{- d(\bm{Q}_t,\bm{\hat{Q}}_t)}$, where $\bm{Q}_t$ is the target orientation at time $t$, $\bm{\hat{Q}}_t$ is the predicted orientation at time $t$, and $d(\bm{Q}_t,\bm{\hat{Q}}_t)$ is the distance between two unit quaternions as given by equation \eqref{eq:dist}.
 
\begin{figure}[!t]
		\centering
		\def\svgwidth{\linewidth}
		{\fontsize{8}{8}
			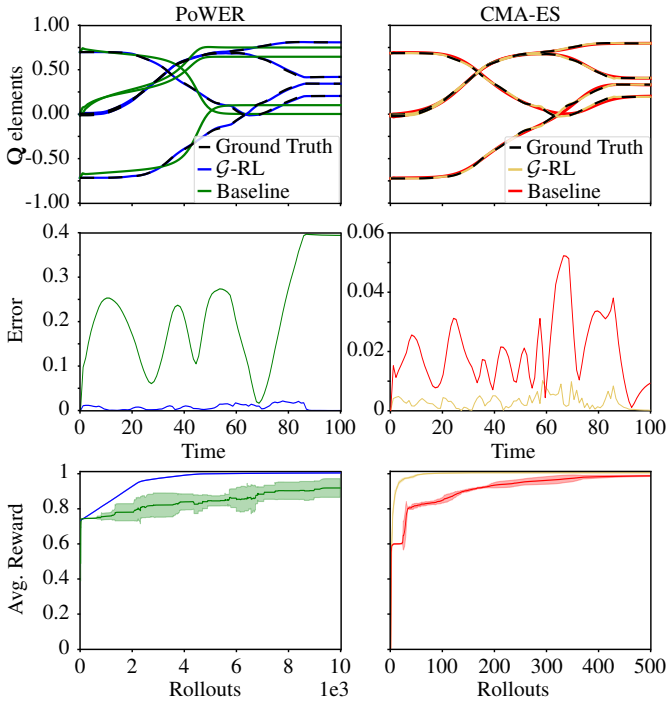}
		\caption{Both \ac{power} (\emph{left}) and \ac{cmaes} (\emph{right})algorithms are used to learn a policy for orientation trajectory tracking, represented as unit quaternions, in a manipulation task. \emph{Top}: The quaternion tracking response of our \ac{grl} approach is compared with the baseline and ground truth. \emph{Middle}: represents the error, computed using~\eqref{eq:dist}, between the trajectories generated by the RL learned policies and the ground truth. \emph{Bottom}: The average reward is plotted with respect to the rollout number, demonstrating the learning progress of the algorithms over time.}
		\label{fig:CMA-ES_Traj_Complex}
\end{figure}

     Fig.~\ref{fig:CMA-ES_Traj_Complex} demonstrates the process of learning a policy for regenerating an orientation trajectory for a specific manipulation task, utilizing both \ac{power} and \ac{cmaes} algorithms. This orientation is denoted by unit quaternions. In the top figure, each unit quaternion is embodied as a 4-dimension vector, each dimension of which records its trajectory through a separate curve. The ultimate policy found using the \ac{power} algorithm is depicted on the top left. The ground truth is represented with a black dashed line, the normalized baseline is a red solid line, and the \ac{grl} is shown as a blue solid line. Alternatively, the top right portrays the optimal policy identified by the \hbox{\ac{cmaes}} algorithm, displaying the ground truth as a black dashed line, the normalized baseline as a red solid line, and the \ac{grl} as a yellow solid line. In the middle, the figure measures the error, determined through the quaternion distance equation ~\eqref{eq:dist}, comparing the divergence between the trajectories produced via the \ac{rl} learned policies and the actual ground-truth. Finally, the figure at the bottom signifies the average reward correlated to the number of rollout trials. We observe that the \ac{grl} error is substantially smaller than the baseline error. When employing the \ac{power} algorithm, the advantage of using \ac{grl} becomes more apparent. However, in both cases, the algorithms' performance significantly improves by applying \ac{grl} compared to the conventional solution, which is the baseline. This shows that utilizing \ac{grl} on a task that involves a trajectory is advantageous because our suggested algorithm predicts the action within the parameterized tangent space and subsequently parallel transport it to the local tangent space that moves over time. This transition within the tangent spaces assures that we optimally utilize Riemannian geometry. It situates the transported action in close vicinity to the origin of the local tangent space (mapped to the local neighborhood of the origin of the tangent space), thereby providing the most suitable configuration.

    \begin{figure}[!t]
		\centering
		\def\svgwidth{\linewidth}
		{\fontsize{8}{8}
			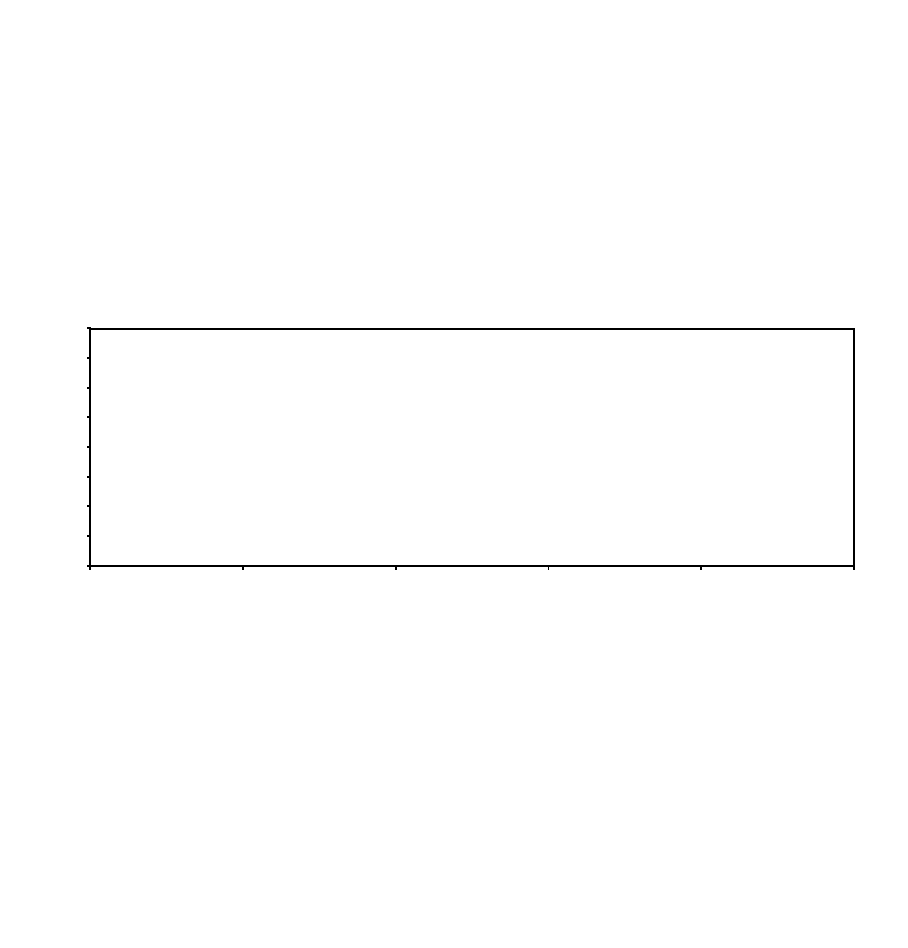}
		\caption{Illustrates the learning of a policy using \ac{cmaes} to regenerate the manipulability ellipsoids, adopted from~\cite{9196952}. Three different approaches are compared: two baselines (Cholesky-based and Mandel-based) and our proposed \ac{grl} approach. \emph{Top}: Show the response of tracking a C-shape trajectory in Cartesian space (black dots). Gray ellipsoids represent the ground truth, Cholesky-based ellipsoids are in green, Mandel-based ellipsoids are in red, and our \ac{grl}-based ellipsoids are in blue. \emph{Middle}: Represents the error, computed by~\eqref{eq:16}, between the trajectories generated by the RL learned policies and the ground truth. \emph{Bottom}: Shows the average reward with respect to the rollout number, indicating the learning progress of the algorithms over time.}
		\label{fig:CMA-ES_Traj_SPD}
\end{figure}

    \begin{figure}[!t]
		\centering
		\def\svgwidth{\linewidth}
		{\fontsize{8}{8}
			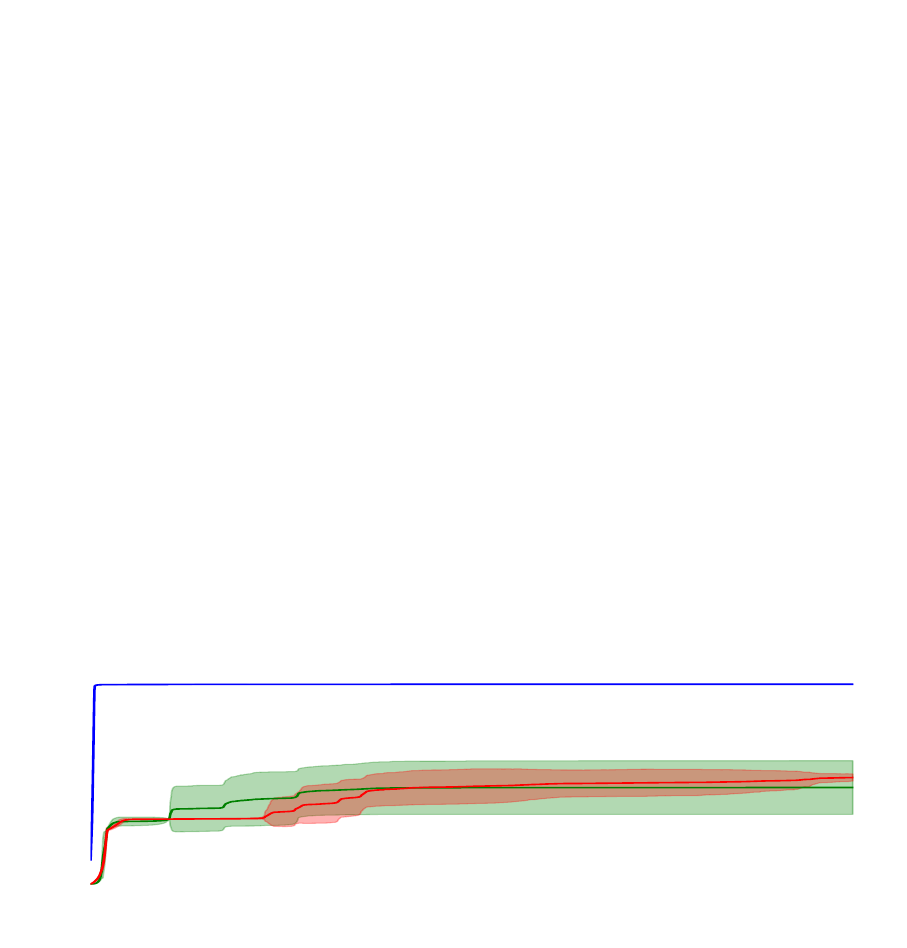}
		\caption{Illustrates the learning of a policy using \ac{power} to regenerate the manipulability ellipsoids, adopted from~\cite{9196952}. Three different approaches are compared: two baselines (Cholesky-based and Mandel-based) and our proposed \ac{grl} approach. \emph{Top}: Show the response of tracking a C-shape trajectory in Cartesian space (black dots). Gray ellipsoids represent the ground truth, Cholesky-based ellipsoids are in green, Mandel-based ellipsoids are in red, and our \ac{grl}-based ellipsoids are in blue. \emph{Middle}: Represents the error, computed by ~\eqref{eq:16}, between the trajectories generated by the \ac{rl} learned policies and the ground truth. \emph{Bottom}: Shows the average reward with respect to the rollout number, indicating the learning progress of the algorithms over time.}
		\label{fig:PoWER_Traj_SPD}
    \end{figure}

    \begin{figure}[!t]
		\centering
		\def\svgwidth{\linewidth}
		{\fontsize{8}{8}
			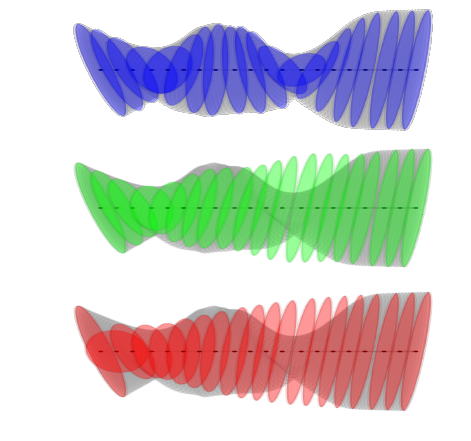}
		\caption{Illustrates learned policies quality using \ac{cmaes} to regenerate the manipulability ellipsoids from~\cite{9196952}, the ground-truth is depicted as the gray ellipsoids, the proposed \ac{grl} approach is depicted as the blue ellipsoids (top) and two baselines (Cholesky is depicted as the green ellipsoids (middle) and Mandel is depicted as the red ellipsoids (bottom)). The shown trajectories are over time.}
		\label{fig:SPD_time}
   \end{figure}

        \subsubsection{\ac{spd} Matrix Trajectory learning problem}\label{subsec:spd_traj_learn}

        As with the Wahba problem, the trajectory learning problem was also replicated using \ac{spd} matrices as well, adjusting the policy to learn a trajectory of \ac{spd} matrices instead of a trajectory of orientations (quaternions).
        
        The reward for the full trajectory in this problem is given by
        $r=\sum_{t=1}^{T}e^{-d({\bm{W}}_t,\widehat{\bm{W}}_t)}$
        where ${\bm{W}}_t$ is the target \ac{spd} matrix at time $t$, $\widehat{\bm{W}}_t$ is the predicted \ac{spd} matrix at time $t$, and $d({\bm{W}}_t,\widehat{\bm{W}}_t)$ is the affine invariant distance between both \ac{spd} matrices given by equation~\eqref{eq:16}.

        In similar context with the orientation trajectory learning problem, and as depicted in Fig.\hbox{~\ref{fig:CMA-ES_Traj_SPD}} and \hbox{~\ref{fig:PoWER_Traj_SPD}} both \hbox{\ac{cmaes}} \hbox{\ac{power}} are used, but the problem here is to regenerate manipulability ellipsoids from [45]. The figures showcases various trajectories: the ground-truth \hbox{\ac{spd}} trajectory illustrated with gray ellipsoids, a Cartesian trajectory represented by black dots, two baseline methods - Cholesky-based \hbox{\ac{spd}} trajectory illustrated with green ellipsoids (left top), Mandel-based \hbox{\ac{spd}} trajectory with red ellipsoids (middle top) - and the proposed \hbox{\ac{grl}} based \hbox{\ac{spd}} trajectory shown with blue ellipsoids (right top). Fig.~\ref{fig:SPD_time} demonstrates the same data with respect to time. Back to Fig.\hbox{~\ref{fig:CMA-ES_Traj_SPD}} and \hbox{~\ref{fig:PoWER_Traj_SPD}}
        The middle of the figure showcases the error between the \hbox{\ac{rl}} learned policies' generated trajectories and the ground-truth values based on the affine invariant distance equation (16). At the bottom portion of the figure, we see an illustration of the average reward in correlation with the rollout number. At first glance at these figures, one can observe that the manipulability ellipsoids generated by \hbox{\ac{grl}} are tracking the ground truth much better than Cholesky and Mandel, and this is quantified by the error figure. Furthermore, the \hbox{\ac{grl}} approach learns faster and converges to a significantly better solution than the commonly used algorithms (Cholesky and Mandel). As we point out about the results of the experiments of the quaternions trajectory learning, applying \hbox{\ac{grl}} on a problem involving a trajectory is most beneficial because our proposed algorithm predicts the action on the parameterization tangent space, then parallel transport it to the local tangent space. This moving tangent space guarantees that we are using the Riemannian geometry in the most appropriate configuration, where the predicted action is located in the close neighborhood of the origin of the local tangent space.

	\subsection{real experiments (Ball-in-a-hole)}
	
	\begin{figure*}[!t]
        \centering
	    \def\svgwidth{\linewidth}
	    {\fontsize{8}{8}
		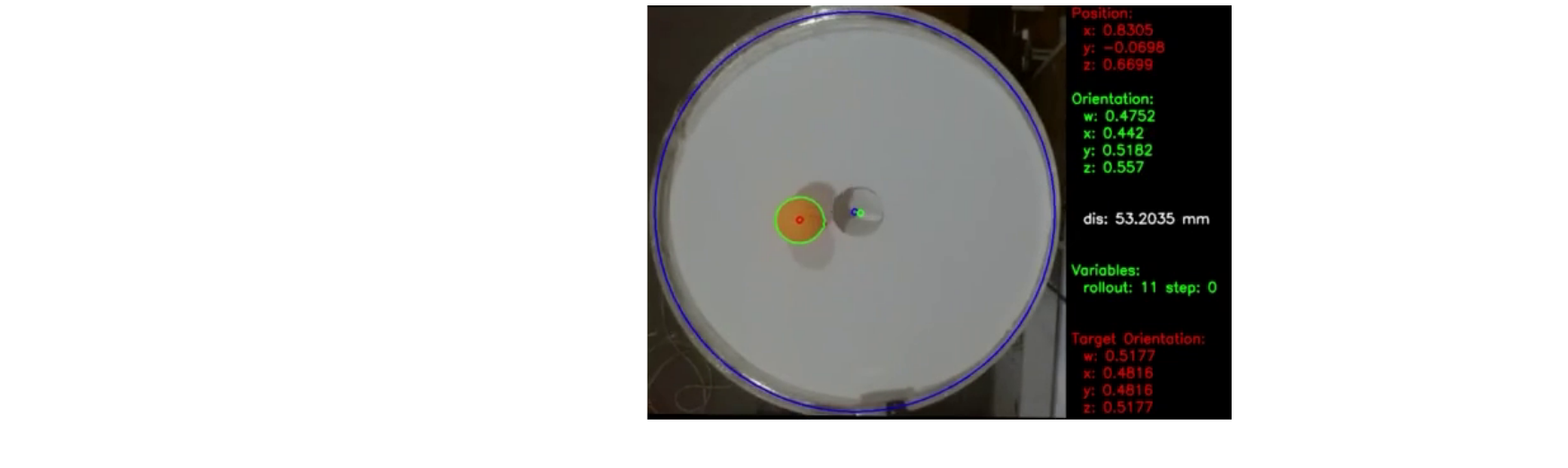}
	    \caption{(a) Ball-in-a-hole problem setup. A plate with a hole in the middle is attached to the robot's end-effector. The plate's circumference is surrounded by cardboard so the ball does not fall outside the plate. A ping pong ball is located on the top of the plate. A camera is also attached to the end-effector in order to measure the distance between the center of the ball and the center of the hole. (b) shows the plate view using the top camera and the data captured from both the vision system and the robot controller.}
	    \label{fig:ball}
        \end{figure*}
        
    \begin{figure}[!t]
        \centering
	    \def\svgwidth{\linewidth}
	    {\fontsize{8}{8}
		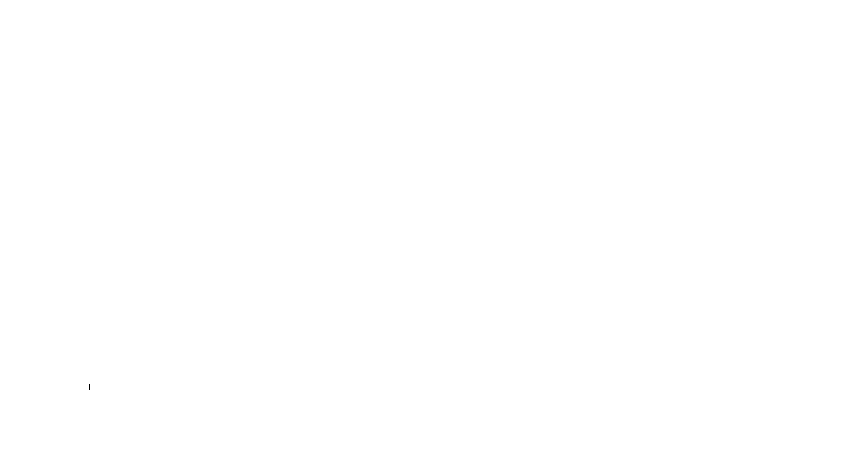}
	    \caption{The expected return of the learned policy in the Ball-in-a-hole evaluation averaged over five runs.}
    	\label{fig:graph}
    \end{figure}
        
    The Ball-in-a-hole problem is a new benchmark proposed in this paper inspired by the Ball-in-a-cup~\cite{sumners1994toys} and the ball balancing~\cite{kopichev2019ball} problems. The problem setup is as depicted in Fig. \ref{fig:ball}, where a plate with a hole in the middle is attached to the end-effector of the TM5-900 \ac{cobot}. A camera is also attached to the robot's end effector and is on a stand to always face the surface of the plate. A ping-pong ball is present on the plate, which has its position tracked by the camera. The robot's end-effector position is fixed, with only its orientation being changed. The state $[\boldsymbol{\mathfrak{s}}_{\mathcal{M}} \dashline \boldsymbol{\mathfrak{s}}_\mathcal{R}]$ includes the current orientation of the end-effector (manifold data $\boldsymbol{\mathfrak{s}}_{\mathcal{M}}$) and the current position of the ball on the plate (Euclidean data $\boldsymbol{\mathfrak{s}}_\mathcal{R}$). On the parameterization tangent space, we concatenate the Euclidean part with the manifold part and deliver it to the policy. The reward is represented by $\exp^{-d}$, where $d$ is the distance between the center of the ball and the center of the hole measured using the vision system. As this is a challenging problem (lightweight ball, noise in the vision system, and with position control), we decided to start each rollout with the ball in the same initial position. 
    
    Regarding the TM5-900 \ac{cobot} limitations, real-time communication is not guaranteed as all communications pass through the TM-Flow software using a \ac{pvt} function. No variable impedance control or admittance control is possible as of writing this paper. Therefore, we had to split the trajectory from one rollout into a number of steps. After each orientation change, the ball's location was immediately read and included in the terminal reward (used to guide the \ac{rl} algorithm). We used the \ac{power} algorithm to learn a policy that moves the ball into the hole, with Fig. \ref{fig:graph} showing the experiment results. The algorithm eventually converged to a local policy, where it learned how to place the ball in the hole via a single axis, as seen in the demonstration video.
    
\section{Discussion}
    As noted in the experimental results (Sec.~\ref{sec:simulations}), as the complexity of the problem increased, the advantage of using \ac{grl} over regular approximation approaches is more significant. This allows us to conclude that in moderately complex problems, the error caused by normalization is significant enough to affect the quality of the solution, and there is a clear advantage in using the proposed \ac{grl}.
    
    As already noted in~\cite{james2022bingham}, \ac{bpp} parameterization relies on the prediction from multiple neural networks, which may introduce significant approximation errors. This culminates in an unstable learning process unlike \ac{gpp} and \ac{grl}. We have experimentally observed this problem of \ac{bpp}, and several attempts were made before representative results were achieved with this approach. On the contrary, the stability of \ac{grl} was on par with \ac{gpp} and better than \ac{bpp}, verifying that the one-to-one mappings between the manifold and tangent space are stable.
    
    We experimentally observed that the average computational overhead of \ac{grl} over \ac{gpp} is about 3\% for \ac{sac} and 6\% for \ac{ppo}. These results were expected as the mappings between the tangent space and manifold are not computationally expensive and it is straightforward to implement. This contrasts with \ac{bpp}, for which we have observed an average overhead of about 33\% for \ac{sac} and 118\% for \ac{ppo}. Moreover, \ac{bpp} involves modifying the distribution and customizing the algorithm to fit. Therefore, we conclude that \ac{grl} can provide a noticeable improvement in solution quality over \ac{gpp} at the cost of a small performance penalty and that it can deliver at least equal results to \ac{bpp} while performing much faster.
    
    Despite the improvement in accuracy, in the case of parameterizing in a single fixed tangent space, \hlnew{where the parameterization and the local tangent spaces to be established at the same fixed point} like in the Wahba problem, it would be most beneficial where data points are in the neighborhood of the origin of the tangent space. This is due to the tangent space projection that locally preserves distances near the origin, while distances measured away from the origin are less accurate. \hlnew{Furthermore, this fixed tangent space should be established on or very close to the mean of the data; otherwise, the algorithm's accuracy can be significantly affected.}
    

    While this work is limited to the $\manS{3}$ and $\manSPD d$ manifolds, it has the potential to be extended to other non-Euclidean manifolds with the proper investigation and analysis. We leave this as future work.
\section{Conclusion}

	Applying \ac{rl} algorithms on geometric data like orientation, manipulability, or stiffness is common in robotics, and these algorithms usually perform better when considering the unique structure of these data. The current study was generally dedicated to this topic and showed how \ac{rl} can be applied to learn geometric actions in the task space (i.e., orientation represented by unit quaternions, and stiffness represented by \mbox{\ac{spd}} matrices); parameterization and optimization are carried on the tangent space, and the policy evaluation is carried on the corresponding manifold $\mathcal{M}$. We found that adapting the Gaussian distribution, which is simple and powerful, to the geometry of non-Euclidean data makes it competitive with alternative distributions (\eg Bingham). Empirical results on both simulation and the physical robot reflect the importance of considering the geometry of non-Euclidean data and how the performance and accuracy of the overall learning process are consequently affected.
	
\section*{Acknowledgment}
The authors would like to thank Ville Kyrki of Aalto University and Luis Figueredo of MIRMI, Technical University of Munich for their help and support in reading and reviewing the math flow of the proposed approach. 
 
\bibliographystyle{IEEEtran}
\bibliography{ref}
	
	\begin{IEEEbiography}[{\includegraphics[width=1in,height=1.25in,clip,keepaspectratio]{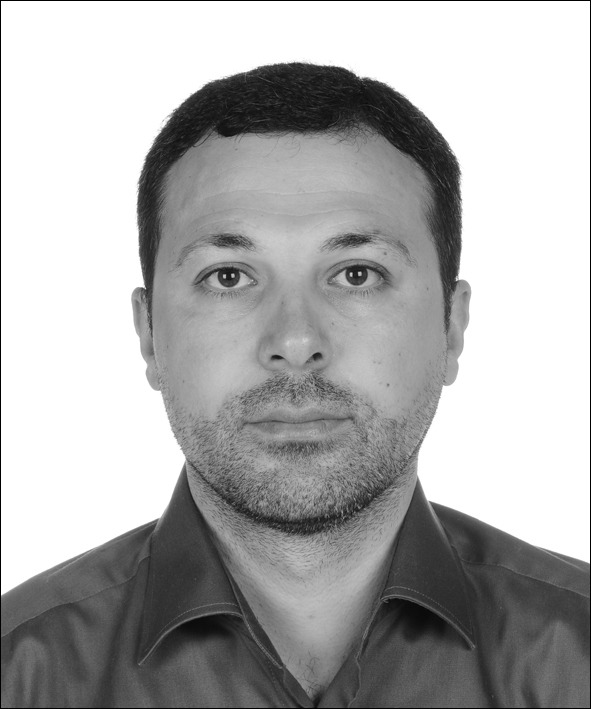}}]{NASEEM Alhousani} received his B.Sc. in computer science and M.Sc. in scientific computing from Birzeit University in 2003 and 2006 respectively. He is currently pursuing a Ph.D. degree in computer engineering at Istanbul Technical University, Istanbul, Turkey. From 2006 to 2015 he worked as a lecturer in the computer science department at Palestine Technical University – Kadoorie. Since 2015 he is a Researcher at ILITRON Energy and Technology, Istanbul, Turkey. His research interest includes reinforcement learning, planning, and learning on Riemannian manifolds.
	\end{IEEEbiography}
	
	\begin{IEEEbiography}[{\includegraphics[width=1in,height=1.25in,clip,keepaspectratio]{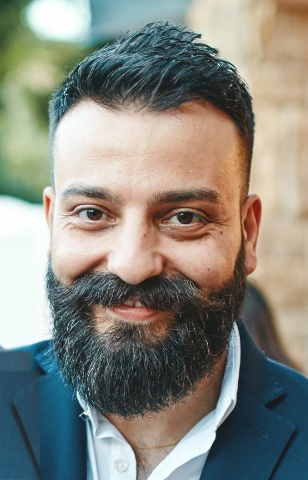}}]{Matteo Saveriano} received his B.Sc. and M.Sc. degree in automatic control engineering from University of Naples, Italy,  in 2008 and 2011, respectively. He received his Ph.D. from the Technical University of Munich in 2017. 	Currently, he is an assistant professor at the Department of Industrial Engineering (DII), University of Trento, Italy. Previously, he was an assistant professor at the University of Innsbruck and a post-doctoral researcher at the German Aerospace Center (DLR). He is an Associate Editor for RA-L and IJRR. His research activities include robot learning, human-robot interaction, understanding and interpreting human activities. Webpage: https://matteosaveriano.weebly.com/ 
	\end{IEEEbiography}
	
	\begin{IEEEbiography}[{\includegraphics[width=1in,height=1.25in,clip,keepaspectratio]{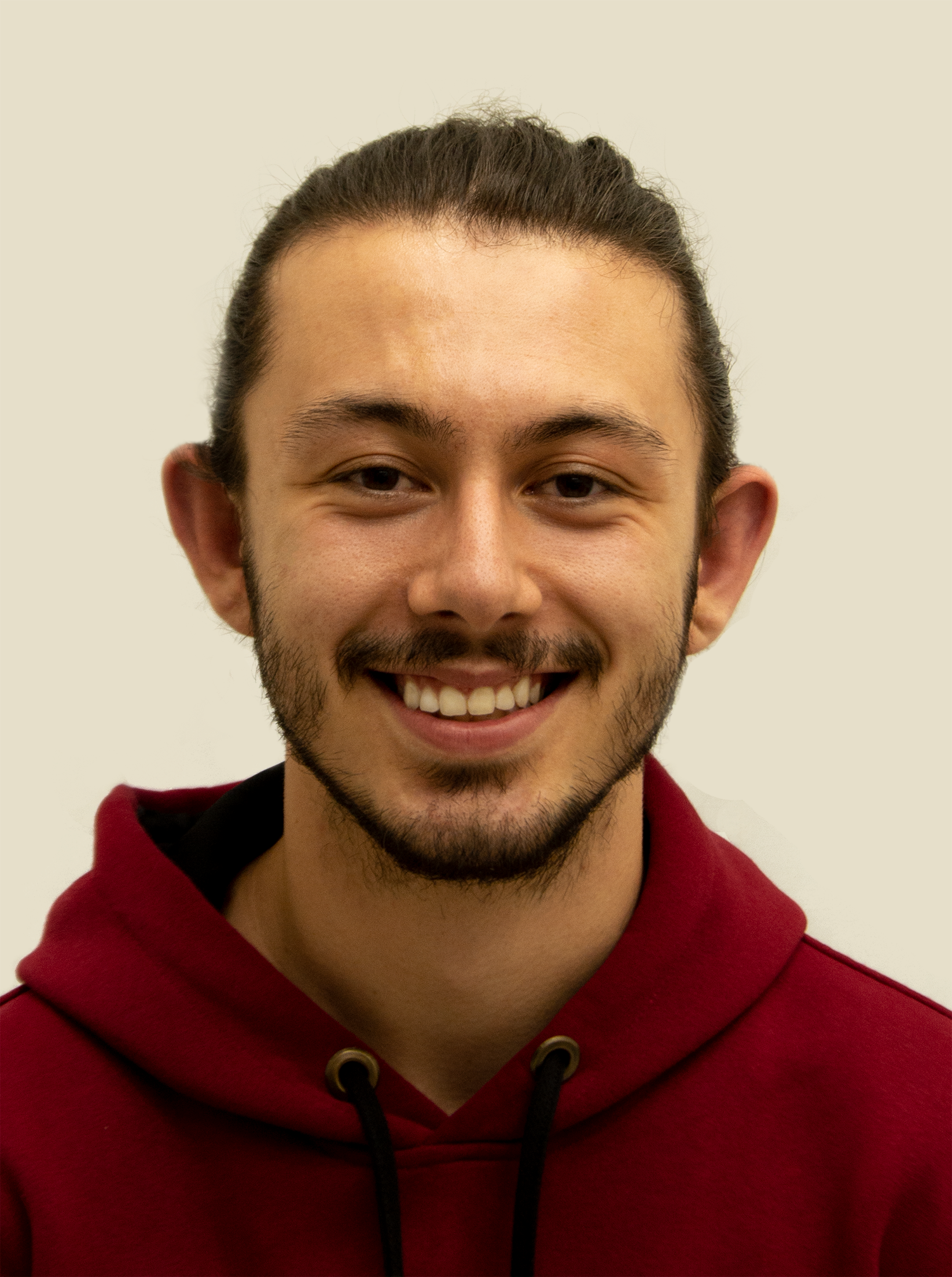}}]{IBRAHIM SEVINC} received his B.Sc. degree in Electronics and Communication Engineering from Istanbul Technical University in 2023. He has been working at MCFLY Robot Technologies, Istanbul, Turkey since 2022.
	\end{IEEEbiography}	
	
	\begin{IEEEbiography}[{\includegraphics[width=1in,height=1.25in,clip,keepaspectratio]{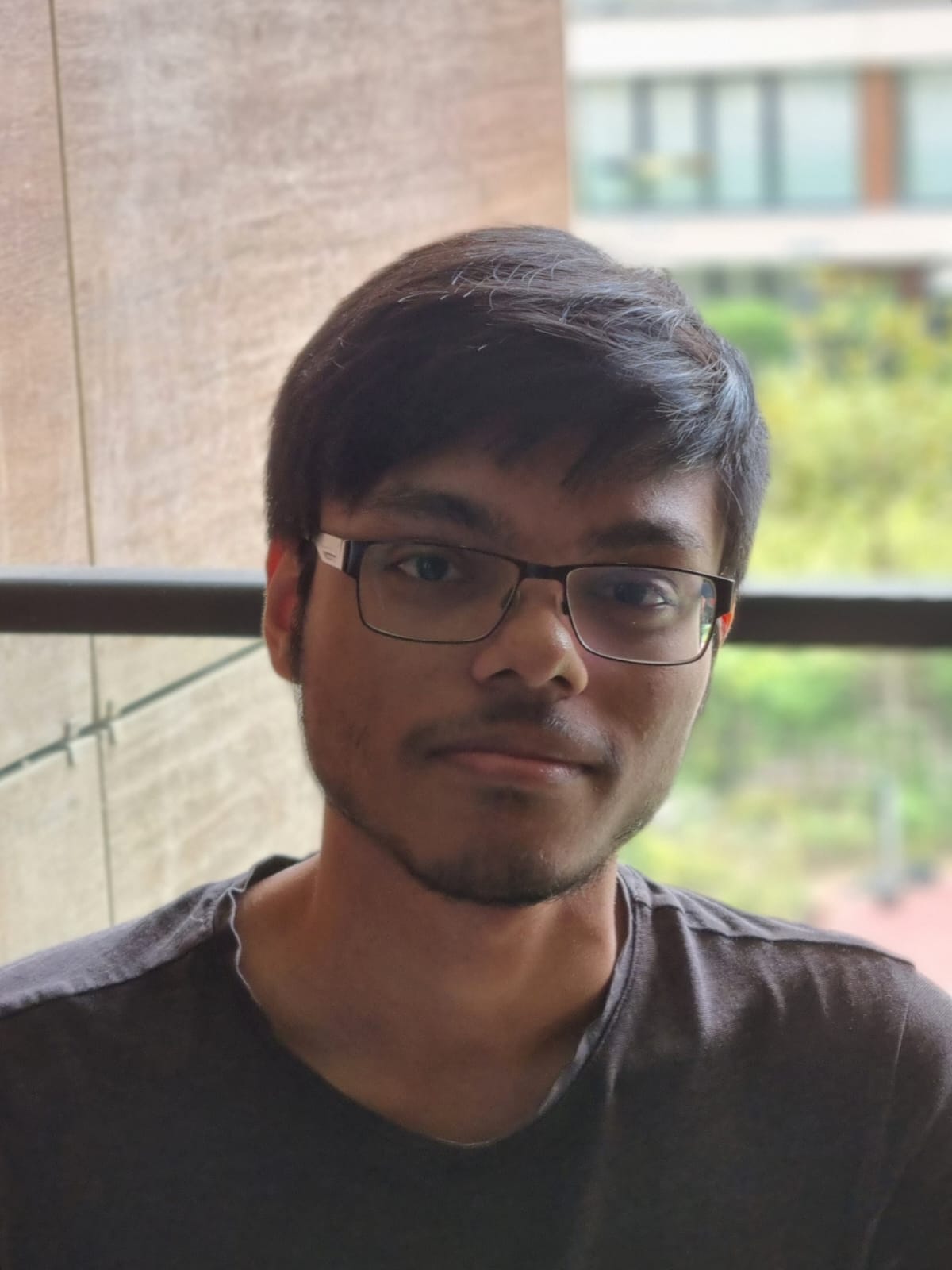}}]{TALHA ABDULKUDDUS} is working towards his B.Sc. Computer Science degree at King's College London, UK. Since 2022, he started working at ILITRON Energy and Information Technologies, Istanbul, Turkey.
	\end{IEEEbiography}		
	
	\begin{IEEEbiography}[{\includegraphics[width=1in,height=1.25in,clip,keepaspectratio]{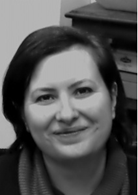}}]{Hatice Kose} is a full Professor at Faculty of Computer and Informatics Engineering, Istanbul Technical University, Turkey, coordinating the GameLab and Cognitive Social Robotics Lab, since 2010. She received her Ph.D. degree from the Computer Engineering Department, Bogazici University, Turkey. From 2006-2010, she worked as a Research Fellow at the University of Hertfordshire. Her current research focuses on gesture communication (involving sign language) and imitation-based interaction games with social humanoid robots for the education and rehabilitation of children with hearing impairment and children with ASD. She is leading several national projects and taking part in several Horiizon2020 projects, Erasmus+ and Cost actions, on social assistive robots, sign language tutoring robots, and human-robot interaction. 
	\end{IEEEbiography}
	
	\begin{IEEEbiography}[{\includegraphics[width=1in,height=1.25in,clip,keepaspectratio]{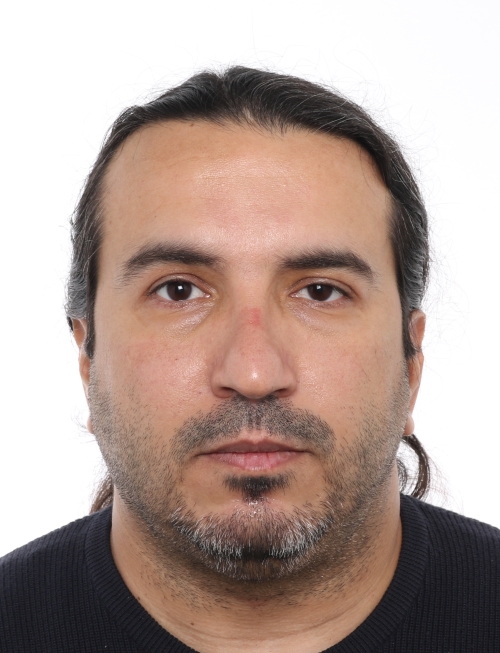}}]{Fares J. Abu-Dakka} received his B.Sc. degree in Mechanical Engineering from Birzeit University, Palestine in 2003 and his DEA and Ph.D. degrees in robotics motion planning from the Polytechnic University of Valencia, Spain in 2006 and 2011, respectively. 

Between 2013 and 2016 he was holding a visiting professor position at ISA of the Carlos III University of Madrid, Spain. In the period between 2016 and 2019, he was a Postdoc at Istituto Italiano di Tecnologia (IIT). During 2019-2022, he was a Research Fellow at Aalto University. Currently, since 2022, he is a Senior Scientist and leading the Robot Learning group at MIRMI, Technical University of Munich, Germany. His research lies in the intersection of control theory, differential geometry, and machine learning, in order to enhance robot manipulation performance and safety.

He is an Associate Editor for ICRA, IROS, and RA-L.  Webpage: https://sites.google.com/view/abudakka/ 
	\end{IEEEbiography}
	
	\EOD
\end{document}